\documentclass[conference]{IEEEtran}
\usepackage{times}

\usepackage[numbers]{natbib}
\usepackage{multicol}
\let\labelindent\relax
\usepackage{enumitem}
\usepackage[bookmarks=true]{hyperref}
\usepackage{booktabs}
\usepackage{graphicx}
\usepackage{caption}
\usepackage{subcaption}
\usepackage{booktabs}
\usepackage{amsmath} 
\usepackage{booktabs}
\usepackage{amsfonts}
\usepackage{tabularx}
\usepackage{siunitx}
\usepackage{makecell}
\usepackage{multirow}
\begin{document}

\title{MOSAIC: Bridging the Sim-to-Real Gap in Generalist Humanoid Motion Tracking and Teleoperation with Rapid Residual Adaptation}

\author{\authorblockN{Zhenguo Sun\authorrefmark{1}$^{1,2}$,
Bo-Sheng Huang\authorrefmark{1}$^{1,3}$,
Yibo Peng\authorrefmark{1}$^{1}$,
Xukun Li\authorrefmark{1}$^{1,2}$,
Jingyu Ma$^{1}$,
Yu Sun$^{1}$,\\ 
Zhe Li$^{1}$,
Haojun Jiang$^{3}$, 
Biao Gao$^{5}$, 
Zhenshan Bing\authorrefmark{2}$^{4}$, 
Xinlong Wang\authorrefmark{2}$^{1}$, 
Alois Knoll$^{2}$}
\authorblockA{$^{1}$Beijing Academy of Artificial Intelligence, 100084 Beijing, China.\\
$^{2}$Technical University of Munich, 85748 Munich, Germany.\\
$^{3}$Tsinghua University, 100084 Beijing, China.\\
$^{4}$Nanjing University, 215163 Suzhou, China.\\
$^{5}$IO-AI.TECH, 518107 Shenzhen, China.
}
\authorblockA{\authorrefmark{1}Equal contribution, \authorrefmark{2}Corresponding author.}}
\makeatletter
    \let\@oldmaketitle\@maketitle%
    \renewcommand{\@maketitle}{
    \@oldmaketitle
    \centering
    \vspace{-2.5pt}
    \includegraphics[width=0.9\textwidth]{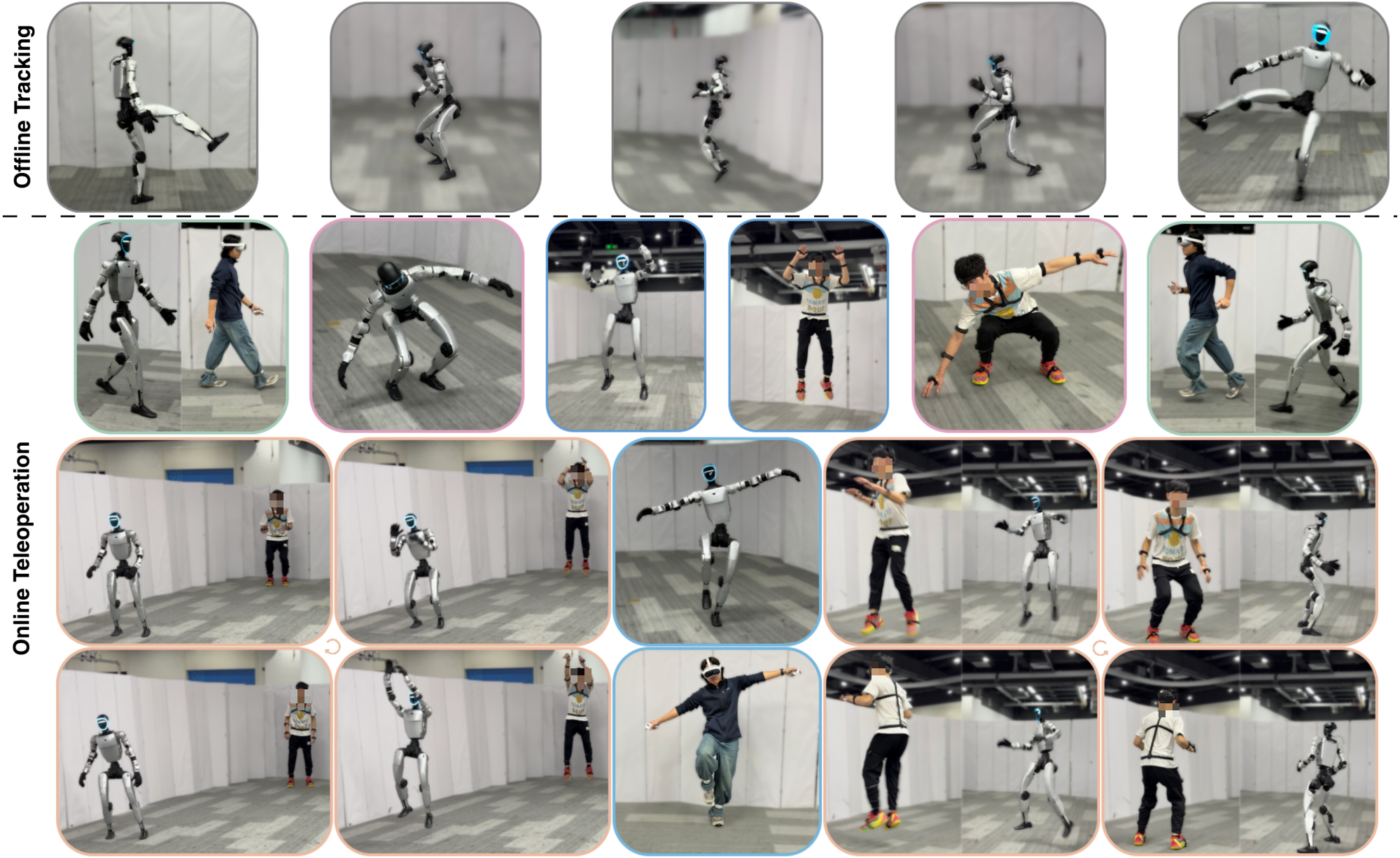}
    \captionof{figure}{
    \textbf{MOSAIC in Action.} MOSAIC enables a single humanoid policy to operate in two modes: offline motion replay (top) and online whole-body teleoperation from multiple wearable interfaces (bottom). In offline replay, the robot robustly tracks diverse and highly dynamic reference motions—walking, running, kicking, kungfu-style strikes, jumping, and squatting. In online teleoperation, MOSAIC faithfully mirrors real-time human motion streams and supports challenging contact-rich and high-agility behaviors, including mid-air jump turns, single-leg support, and jump-shot–style movements.
    }
    \label{fig:teaser}
    \vspace{-9.5pt}
    \addtocounter{figure}{-1}%
    }
\makeatother

\maketitle

\begin{abstract}
Generalist humanoid motion trackers have recently achieved strong simulation metrics by scaling data and training, yet often remain brittle on hardware during sustained teleoperation due to interface- and dynamics-induced errors. We present MOSAIC, an open-source, full-stack system for humanoid motion tracking and whole-body teleoperation across multiple interfaces. MOSAIC first learns a teleoperation-oriented general motion tracker via RL on a multi-source motion bank with adaptive resampling and rewards that emphasize world-frame motion consistency, which is critical for mobile teleoperation. To bridge the sim-to-real interface gap without sacrificing generality, MOSAIC then performs rapid residual adaptation: an interface-specific policy is trained using minimal interface-specific data, and then distilled into the general tracker through an additive residual module, outperforming naive fine-tuning or continual learning. We validate MOSAIC with systematic ablations, out-of-distribution benchmarking, and real-robot experiments demonstrating robust offline motion replay and online long-horizon teleoperation under realistic latency and noise. Project page: \href{https://baai-humanoid.github.io/MOSAIC/}{baai-humanoid.github.io/MOSAIC}.
\end{abstract}

\IEEEpeerreviewmaketitle

\section{Introduction}
\label{sec:intro}

Humanoid motion imitation has advanced rapidly since DeepMimic~\cite{peng2018deepmimic}, turning whole-body control into a scalable learning problem. Early reinforcement learning (RL) imitation systems achieved strong tracking for single clips or small motion sets~\cite{he2025asap,luo2023phc}, and extended to more dynamic, contact-rich behaviors~\cite{xie2025kungfubot,liao2025beyondmimic}. More recently, the focus has shifted toward general motion tracking---a single policy trained on large motion corpora (e.g., GMT~\cite{chen2025gmt}, Any2Track~\cite{zhang2025any2track}, UniTracker~\cite{yin2025unitracker}, KungfuBot2~\cite{han2025kungfubot2})---alongside end-to-end teleoperation systems for scalable whole-body data collection and mobile manipulation (e.g., ExBody~\cite{ji2024exbody2}, OmniH2O~\cite{he2024omnih2o}, HumanPlus~\cite{fu2024humanplus}, CLONE~\cite{li2025clone}, TWIST~\cite{ze2025twist, ze2025twist2}, SONIC~\cite{luo2025sonic}). Together, these directions suggest a clear trajectory: as tracking becomes more general and reliable, whole-body teleoperation becomes a foundational capability for both remote operation and large-scale demonstration collection.

Yet, strong simulation tracking metrics do not reliably translate to hardware performance, especially for long-horizon locomotion and sustained teleoperation. In our experience, a tracker can reach state-of-the-art (SOTA) offline scores while still stumbling, losing contact stability, or drifting in world-frame motion on a real robot. This metric--deployment gap is amplified by what enables generality in the first place: heterogeneous, fragmented motion data and multiple teleoperation interfaces (inertial MoCap suits, VR trackers, etc.) with distinct latency, noise, and retargeting biases. Because humanoid locomotion couples underactuated base dynamics with repeated contacts, small interface-induced errors can compound into long-horizon failures, making “accurate tracking” insufficient for “usable teleoperation.”

We present \textbf{MOSAIC} (\textbf{MO}tion tracking \textbf{S}ystem with \textbf{A}daptive \textbf{I}nterface \textbf{C}orrection), a deployable whole-body motion tracking and teleoperation system designed to close this gap. 
MOSAIC has two core components. 
First, we train a teleoperation-oriented general motion tracking policy via single-stage RL, using multi-source dataset with an adaptive resampling strategy, and a reward design that improves full-body pose tracking while explicitly emphasizing global motion consistency that matters for teleoperation (e.g., tracking body positions and world-frame motion). 
Second, we introduce a small-data interface adaptor that enables rapid adaptation to new teleoperation interfaces using a short calibration session within only a few additional motion data(e.g. $\sim$30 minutes). 
Rather than continual fine-tuning of the general tracker---which we find can underperform due to imbalance and generality degradation---we train an adaptation policy on a small dataset and distill its improvement into the general tracker through a residual correction module, analogous in spirit to plug-in adaptation: it preserves broad motion coverage while injecting interface-specific corrections that substantially improve real-robot stability. 
Empirically, this residual adaptation yields more direct gains than adding the same data for continual learning, and also improves deployment more reliably than periodic-motion augmentation \cite{li2024fld}.

To make MOSAIC practical across robots and software stacks, we further develop \textbf{RobotBridge}, a modular deployment framework that standardizes the interfaces among motion references, policy execution, simulator/robot backends, and low-level controllers. 
RobotBridge supports consistent evaluation from simulation to real robots and enables portability across humanoid platforms with minimal changes, improving reproducibility and facilitating fair comparisons. 
Finally, we complement the system with a multi-source motion dataset and a full open-source release of training, deployment, checkpoints, data, and data collection pipelines, together with extensive ablations and real-robot studies that turn our engineering iterations into actionable insights for the community.

\noindent\textbf{Contributions.} Our contributions are summarized as follows:
\begin{itemize}[leftmargin=1.2em]
    \item \textbf{MOSAIC: a deployable tracker \& teleoperation system.} A unified whole-body motion tracking stack supporting offline motion replay and online teleoperation from multiple interfaces, achieving strong tracking accuracy and substantially improved real-robot robustness.
    \item \textbf{Teleoperation-oriented general tracker learning.} A RL formulation with a reward design tailored to teleoperation, explicitly improving global motion consistency.
    \item \textbf{Residual adaptation.} A plug-in residual distillation mechanism that adapts a general tracker to a new teleoperation interface using small data, improving teleoperation stability and the ability to support long-horizon loco-manipulation tasks while preserving the tracker’s motion generality.
    \item \textbf{RobotBridge and open resources.} A modular deployment framework connecting policies, simulators, robot SDKs, and controllers, along with a high-quality motion dataset, trained checkpoints, and an open-source release of training/deployment/data-collection pipelines.
\end{itemize}
\section{Related Work}
\label{sec:related}

\textbf{From motion imitation to generalist tracking.}
RL-based imitation progressed from tracking single or small motion sets toward general motion tracking policies trained on large corpora. Recent generalist trackers emphasize scaling data/model capacity and balancing heterogeneous motion distributions. GMT~\cite{chen2025gmt} couples adaptive sampling with a motion MoE (Mixture of Experts) to improve coverage and specialization across diverse motions.
Any2Track~\cite{zhang2025any2track} highlights real-world robustness under disturbances via explicit adaptation. UniTracker~\cite{yin2025unitracker} uses a teacher--student pipeline with a learned latent representation, and KungfuBot2~\cite{han2025kungfubot2} underscores minute-scale stability as a persistent bottleneck. Different from works that primarily optimize tracking metrics or disturbance adaptation in isolation, we emphasize a teleoperation-ready system perspective: stable minute-scale tracking on a real humanoid together with a reproducible training recipe that remains effective under heterogeneous, uneven-length motion corpora.

\textbf{Whole-body teleoperation systems.}
In parallel, teleoperation systems prioritize end-to-end usability---latency tolerance, responsiveness, robustness, and scalable data collection---as a prerequisite for whole-body mobile manipulation.
TWIST~\cite{ze2025twist} shows that combining imitation objectives with policy learning can yield effective whole-body teleoperation, while TWIST2~\cite{ze2025twist2} further improves portability and data scalability by relaxing reliance on specialized capture setups.
SONIC~\cite{luo2025sonic} advocates a ``from scale tracking to generalist control'' perspective and supports multiple motion input interfaces, highlighting the importance of bridging tracking foundations with deployable systems.
In contrast to teleop works that center on a specific capture interface or data collection setup, we unify offline replay and online IMU teleoperation under one motion tracker, and we explicitly address long-horizon stability and dataset heterogeneity through our multi-source pipeline and adaptive resampling—packaged for open release.

\textbf{Residual adaptation.}
A recurring lesson in real-robot deployment is that a strong base controller can remain brittle when the input interface or deployment distribution shifts, motivating lightweight adaptation on top of a frozen backbone.
Residual learning is a classical strategy to refine an existing controller/policy with an additive corrective module~\cite{johannink2019residual}, and has been repeatedly validated as a practical way to improve real-world performance without retraining from scratch.
Recent humanoid work such as ResMimic~\cite{zhao2025resmimic} adopts a two-stage ``general motion prior + residual refinement'' recipe, while adaptation-module paradigms such as RMA~\cite{kumar2021rma} highlight the effectiveness of separating a robust base policy from a small, fast adaptation component.
Complementary to these, adapter-based parameter-efficient tuning has been explored for robot policies to transfer skills across embodiments while updating only small modules~\cite{lu2024learning}.
Different from residual/adaptation works that primarily target task-specific manipulation or dynamics changes, we emphasize system stability and generality, demonstrate minute-scale real-humanoid results under out-of-distributions (OOD) setting.

\textbf{Multi-source motion data and reusable priors.}
Large-scale motion corpora such as AMASS \cite{mahmood2019amass} provide diverse human motion in a unified format and have become a backbone for learning motion priors.
Meanwhile, modern motion generation models (e.g., GENMO \cite{li2025genmo}) enable scalable synthesis and augmentation of motion distributions, which is increasingly used to expand coverage beyond captured MoCap data. At a higher level, Behavior Foundation Models and recent surveys frame whole-body control as learning reusable priors through large-scale pretraining and distillation for broad downstream use \cite{zeng2025bfm, yuan2025survey}. 
However, translating heterogeneous and imperfect motion sources into a teleoperation-ready policy remains non-trivial, because dataset imbalance, short clips, and interface-specific noise can yield strong metrics but weak hardware performance.
Our contribution fits this foundation/pretraining trajectory but targets a concrete systems gap: we combine multi-source data unification with a two-level adaptive resampling mechanism and stage-wise training so that heterogeneous datasets translate into a stable, deployable motion-tracking/teleop controller—validated with minute-scale real-robot tracking and released as an end-to-end recipe.
\section{System Overview}
\label{sec:sys}

\setcounter{figure}{1}
\begin{figure*}[htbp]
    \centering
    \includegraphics[width=0.95\linewidth]{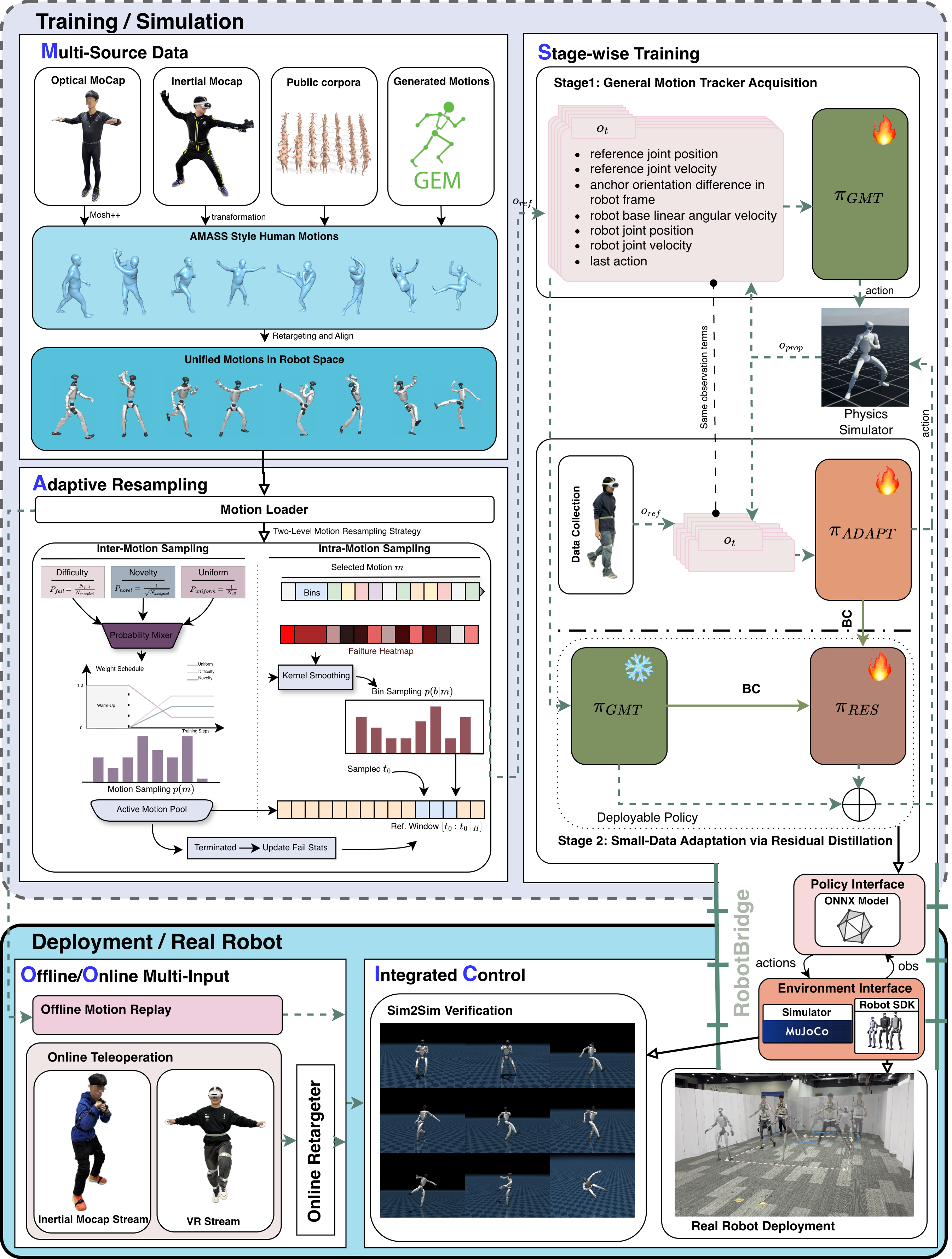}
    \caption{\textbf{MOSAIC System Overview.} MOSAIC consists of a unified training–deployment pipeline for humanoid motion tracking and teleoperation. Training/Simulation aggregates heterogeneous multi-source motions, two-level adaptive resampling, policy training process, yielding a deployable policy that preserves generality while improving real-robot robustness. Deployment/Real Robot supports both offline motion replay and online teleoperation. Finally, RobotBridge provides a modular interface that enables consistent evaluation and portable deployment across platforms.}
    \label{fig:system_overview}
\end{figure*}

MOSAIC is a teleoperation-ready humanoid motion tracker that executes a single learned policy in two modes: (i) offline motion replay (tracking a stored reference clip), and (ii) online teleoperation (tracking a live human motion stream from wearable interfaces such as inertial MoCap or VR trackers). Figure~\ref{fig:system_overview} summarizes the pipeline. 

\subsection{Inputs: One-Step Reference}
Both modes share the same policy interface: at each control step the policy consumes only the \textbf{next reference frame} (one-step lookahead), rather than a multi-step future window. In offline replay, the next frame is queried directly from the stored robot-space clip; in online teleoperation, human measurements are \emph{retargeted online} into the same robot-space format (implemented using a GMR-style retargeting pipeline~\cite{joao2025gmr,ze2025gmr}).

\subsection{Outputs and Control Execution}
The policy outputs joint position targets $q^{des}_t \in \mathbb{R}^{29}$ at 50\,Hz, which are executed by a joint-space PD controller:
\begin{equation}
\tau_t = K_p(q^{des}_t - q_t) + K_d(\dot{q}^{des}_t - \dot{q}_t),
\end{equation}
with safety handling (torque/velocity saturation and joint limits). The same control stack is used in simulation and on robot.


\section{Multi-Source Data \& Adaptive Resampling}
\label{sec:resampling}

MOSAIC is trained on a heterogeneous multi-source motion corpus spanning markedly different distributions (capture devices, demonstrators, noise/latency profiles, and clip lengths). 
In total, we use approximately 64 hours of motion data and will release all collected data together with preprocessing scripts and prompts/licenses where applicable.

\subsection{Multi-Source Motion Data}
Our corpus combines: (i) in-house optical MoCap, (ii) in-house inertial MoCap, (iii) public motion datasets (AMASS~\cite{mahmood2019amass} and OMOMO~\cite{li2023omomo}), (iv) generated motions via GENMO~\cite{li2025genmo}, and (v) a small amount of interface-specific teleoperation data used for rapid adaptation, as listed in Table~\ref{tab:data_source_summary}.
All sources are converted into unified robot-space representation at a fixed rate (50\,Hz) following BeyondMimic~\cite{liao2025beyondmimic}.

\begin{table}[t]
    \centering
    \small
    \setlength{\tabcolsep}{4pt}
    \renewcommand{\arraystretch}{1.2}
    \begin{tabular}{lccc}
    \toprule
    \textbf{Source} & \textbf{Scale} & \textbf{Avg. length} & \textbf{Redistribution} \\
    \midrule
    Optical MoCap & 3.1h & medium & Yes (ours) \\
    Inertial MoCap & 7.0h & med/long & Yes (ours) \\
    Public Corpora & 51.0h & varied & scripts only \\
    GENMO Motions & 2.2h & short & prompt \& motions \\
    Adaptation Motions & 1.0h & short & Yes (ours) \\
    \bottomrule
    \end{tabular}
    \caption{\textbf{Dataset summary.} We will open-source all high-quality multi-source motion data collected in-house.}
    \label{tab:data_source_summary}
\end{table}

\subsection{Adaptive resampling}
We employ adaptive resampling at two levels to handle uneven lengths and difficulty without manual dataset balancing.
\emph{(i) Within-motion sampling} follows BeyondMimic~\cite{liao2025beyondmimic}: each motion is divided into coarse time bins, which are sampled from a failure-aware distribution updated online, focusing learning on unstable segments while maintaining coverage.
\emph{(ii) Motion-level sampling} periodically assigns environments to motions using a simple mixture of difficulty (motions with higher failure rates), novelty/coverage (less-assigned motions), and a uniform term to prevent collapse.
Together, these two samplers stabilize training over heterogeneous sources and clip-length regimes while preserving broad motion diversity.
\section{Policy Training}
\label{sec:method}

\subsection{Motion Tracking Policy Learning}
\label{sec:method:policy}

We formulate the motion tracking task as a partially observable Markov decision process
(POMDP) defined by $(\mathcal{S}, \mathcal{O}, \mathcal{A}, P, R, \gamma)$, where 
$\mathcal{S}$ denotes the full environment state space and $\mathcal{O}$ is
the observation space available to the robot. At each time
step $t$, the robot receives an observation $\mathbf{o}_t \in \mathcal{O}$ and produces an action
$\mathbf{a}_t \in \mathcal{A}$ through the policy $\pi(\mathbf{a}_t \mid \mathbf{o}_t)$. The
environment evolves with stochastic dynamics $P(\mathbf{s}_{t+1} \mid \mathbf{s}_t, \mathbf{a}_t)$,
and the robot receives a scalar reward $r_t = R(\mathbf{s}_t, \mathbf{a}_t)$ based on
motion-tracking fidelity and teleoperation objectives.  

We adopt the asymmetric Proximal Policy Optimization (PPO) algorithm \cite{schulman2017ppo} to train two motion tracking policies, a general motion
tracking policy $\pi_{\mathrm{GMT}}$ and an adaptation motion tracking policy $\pi_{\mathrm{ADAPT}}$, each handling
different aspects of whole-body control and real-world teleoperation. 
We employ an asymmetric actor–critic formulation, where the critic has access to privileged state information during training. Observation terms, and reward terms and its weights, as well as domain randomization are summarized in
Tables~\ref{tab:obs_noise}, \ref{tab:rewards}; domain randomization applied during policy training is mainly followed BeyondMimic~\cite{liao2025beyondmimic}, including body friction, torso center-of-mass offset, and base velocity perturbation, etc.

\paragraph{General Motion Tracking Policy}  
The general motion tracking policy $\pi_{\mathrm{GMT}}$ is trained on heterogeneous multi-source
motion datasets and augmented with world-frame teleoperation tracking rewards that emphasize
kinematic fidelity while imposing global trajectory constraints. Although $\pi_{\mathrm{GMT}}$
achieves high fidelity for local whole-body motions, it lacks robustness for long-horizon locomotion.
Humanoid locomotion involves high-frequency contact transitions and high-dimensional and non-smooth systems with many physical constraints~\cite{hwangbo2019agile}, requiring
the policy to cover a wide repertoire of motion patterns. Multi-source datasets often provide
incomplete coverage, and different teleoperation devices induce gait variations even for the same
motion type. As a result, $\pi_{\mathrm{GMT}}$ struggles to generalize to unseen contact-rich
interactions and varying gait transitions, leading to stability degradation and tracking errors
during long-horizon deployment.

\paragraph{Adaptation Motion Tracking Policy}  
To bridge the domain gap between heterogeneous motion datasets and teleoperation signals,
we introduce an adaptation policy $\pi_{\mathrm{ADAPT}}$, trained on a small adaptation dataset collected
using teleoperation devices (e.g., VR tracker). This adaptation motion tracking policy captures intrinsic noise characteristics
and device-specific gait variations. $\pi_{\mathrm{ADAPT}}$ uses the same observation and reward as $\pi_{\mathrm{GMT}}$, while adaptation
data enable it to specialize in the control patterns associated with specific hardware interfaces,
fitting the localized motion patterns of the adaptation data rather than the broad behavioral
repertoire of the original multi-source dataset.

\subsection{Motion-Conditioned Residual Adaptor Distillation}
\label{sec:method:residual}

Although $\pi_{\mathrm{GMT}}$ mastered a diverse range of foundational motions, direct fine-tuning to enhance long-horizon stability or adapt to new teleoperation interfaces is often counterproductive, as it risks catastrophic forgetting of previously learned behaviors. To address this, we introduce a residual adapter $\pi_{\mathrm{RES}}$ that outputs an action refinement within the same action space. The student policy $\pi_S$ is thus defined as:
\begin{equation}
    \pi_S(\mathbf{o}_t) = \pi_{\mathrm{GMT}}(\mathbf{o}_t) + \pi_{\mathrm{RES}}(\mathbf{o}_t),
\end{equation}
where the parameters of the base policy $\pi_{\mathrm{GMT}}$ remain frozen throughout the residual training stage.

\paragraph{Zero-biased Residual Initialization} 
The objective of residual learning is to perform \emph{sparse adjustments} atop a robust frozen backbone rather than re-learning the full control manifold. Following ResMimic~\cite{zhao2025resmimic}, we initialize the final layer of $\pi_{\mathrm{RES}}$ with a near-zero weight gain and zero bias. This keeps the initial residual output close to zero, making early training updates conservative and preventing the sudden degradation of the base policy's performance.

\paragraph{Dual-Teacher Behavior Cloning} 
The student policy is trained through a multi-teacher distillation framework. During training, each rollout transition is categorized by its motion regime: teleoperation adaptation data ($\mathcal{D}_{\mathrm{adapt}}$) or general heterogeneous data ($\mathcal{D}_{\mathrm{gmt}}$). The distillation loss minimizes the mean squared error (MSE) between the composite student policy and the corresponding teacher actions:
\begin{equation}
    \mathcal{L}_{\mathrm{distill}} = \sum_{k \in \{ \text{ADAPT, GMT} \}} w_k \mathbb{E} \left[ \left\| \pi_S(\mathbf{o}_t) - \pi_{(k)}(\mathbf{o}_t) \right\|_2^2 \right],
\end{equation}
where $w_k$ denotes the weighting factor for each regime. Here, $\pi_{(k)}$ serves as the teacher policy for its respective domain: $\pi_{\mathrm{GMT}}$ provides target actions from the foundation model to maintain general proficiency, while $\pi_{\mathrm{ADAPT}}$ provides expert demonstrations for the specific adaptation tasks. 

By optimizing this delta action by dual teacher respectively, the residual component $\pi_{\mathrm{RES}}$ learns to effectively compensate for domain gaps and teleoperation biases while preserving the foundational skills inherent in $\pi_{\mathrm{GMT}}$.


\begin{table}[t]
    \centering
    \small
    \setlength{\tabcolsep}{6pt}
    \renewcommand{\arraystretch}{1.2}
    \begin{tabular}{l l c}
    \toprule
    \textbf{Category} & \textbf{Observation Term} & \textbf{Noise} \\
    \midrule
    \multirow{7}{*}{Prop.}
    & reference joint position                 & --              \\
    & reference joint velocity                 & --              \\
    & anchor orientation (robot frame) & $\mathcal{U}(-0.05,\,0.05)$ \\
    & robot base angular velocity                  & $\mathcal{U}(-0.2,\,0.2)$   \\
    & robot joint position              & $\mathcal{U}(-0.01,\,0.01)$ \\
    & robot joint velocity              & $\mathcal{U}(-0.5,\,0.5)$   \\
    & last action                           & -- \\
    \midrule
    \multirow{5}{*}{Priv.}
    & anchor position (robot frame)    & -- \\
    & body position (robot frame)             & -- \\
    & body orientation (robot frame)          & -- \\
    & robot base linear velocity                   & -- \\
    & reference base linear velocity         & -- \\
    \bottomrule
    \end{tabular}
    \caption{\textbf{Observation terms and noises used during policy training.}
    The actor uses the \textit{Prop} observation stack with a 5-step history, while the critic uses the full set of observations including \textit{Prop} and \textit{Priv} terms with current step.}
    \label{tab:obs_noise}
\end{table}

\begin{table}[t]
\centering
\small
\setlength{\tabcolsep}{6pt}
\renewcommand{\arraystretch}{1.3}
\begin{tabular}{l c l}
\toprule
\textbf{Reward Term} & \textbf{Weight} & \textbf{Parameters} \\
\midrule
\multicolumn{3}{l}{\textit{Tracking rewards}} \\
\addlinespace[0.1cm]
Motion global anchor position & +0.5 & std=0.3 \\
Motion global anchor orientation & +0.5 & std=0.4 \\
Motion body position & +1.0 & std=0.3 \\
Motion body orientation & +1.0 & std=0.4 \\
Motion body linear velocity & +1.5 & std=1.0 \\
Motion body angular velocity & +1.5 & std=3.14 \\
Motion anchor linear velocity & +1.0 & std=1.0 \\
\midrule
\addlinespace[0.1cm]
\multicolumn{3}{l}{\textit{Teleoperation rewards}} \\
\addlinespace[0.1cm]
Motion global body position & +1.0 & std=0.5 \\
Motion global VR & +0.5 & std=0.5 \\
Motion global feet position & +1.0 & std=0.5 \\
Motion global body orientation & +0.5 & std=0.5 \\
Motion global body angular velocity & +0.5 & std=3.14 \\
Motion global body linear velocity & +0.5 & std=1.0 \\
\midrule
\addlinespace[0.1cm]
\multicolumn{3}{l}{\textit{Penalty terms}} \\
\addlinespace[0.1cm]
Undesired contacts & -0.05 & threshold=1.0N \\
Action rate L2 & -0.1 & -- \\
Joint limit & -10.0 & -- \\
Joint acceleration & -2.5e-7 & -- \\
Joint torque & -1e-5 & -- \\
\bottomrule
\end{tabular}
\caption{\textbf{Reward terms and weights.} Our design combines the tracking objectives from BeyondMimic~\cite{liao2025beyondmimic} with additional global teleoperation-oriented terms inspired by prior motion-imitation systems such as ASAP~\cite{he2025asap} and KungfuBot~\cite{xie2025kungfubot}, while regularizing contacts, action smoothness, and joint limits/torques to improve stability and deployability.}
\label{tab:rewards}
\end{table}

\begin{figure}[htbp]
    \centering
    \begin{subfigure}{\columnwidth}
        \centering
        \includegraphics[width=1.0\linewidth]{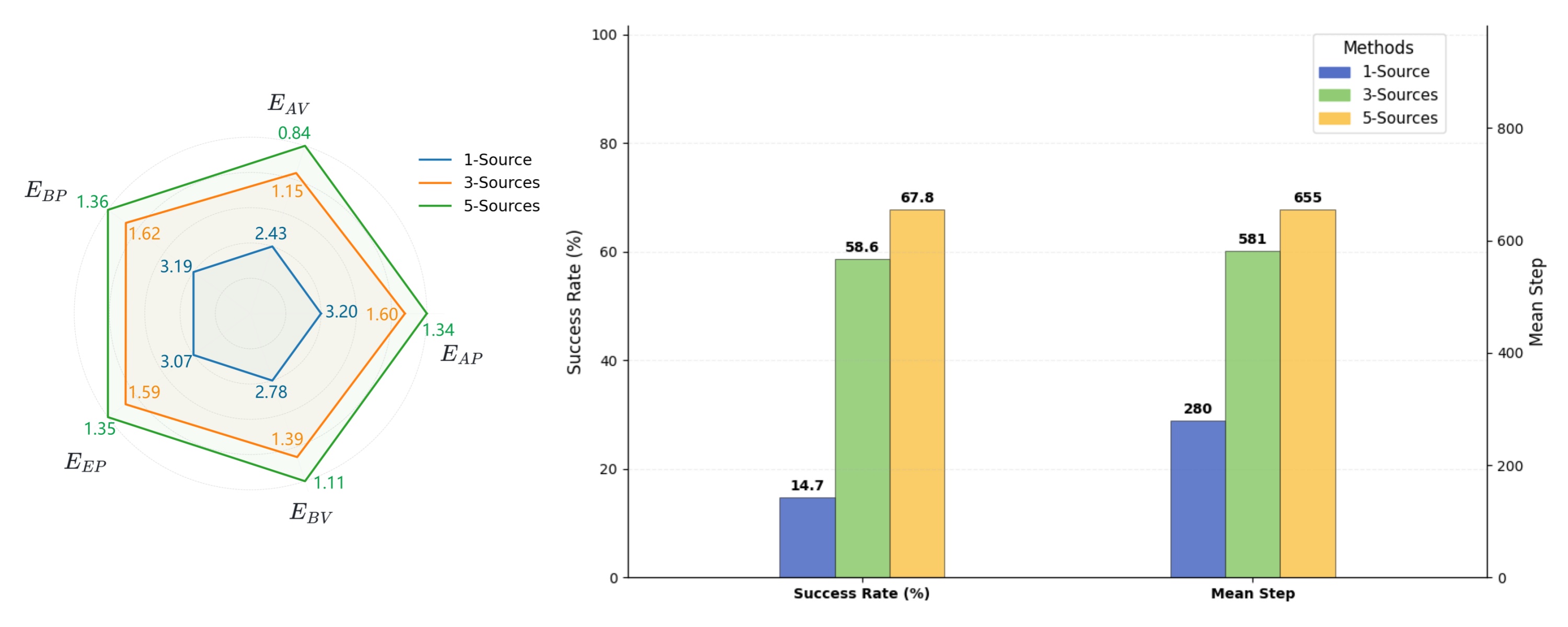}
        \caption{Quantitative Comparison of Data Source.}
        \label{fig:compare_multi_source}
    \end{subfigure}
    
    \begin{subfigure}{\columnwidth}
        \centering
        \includegraphics[width=1.0\linewidth]{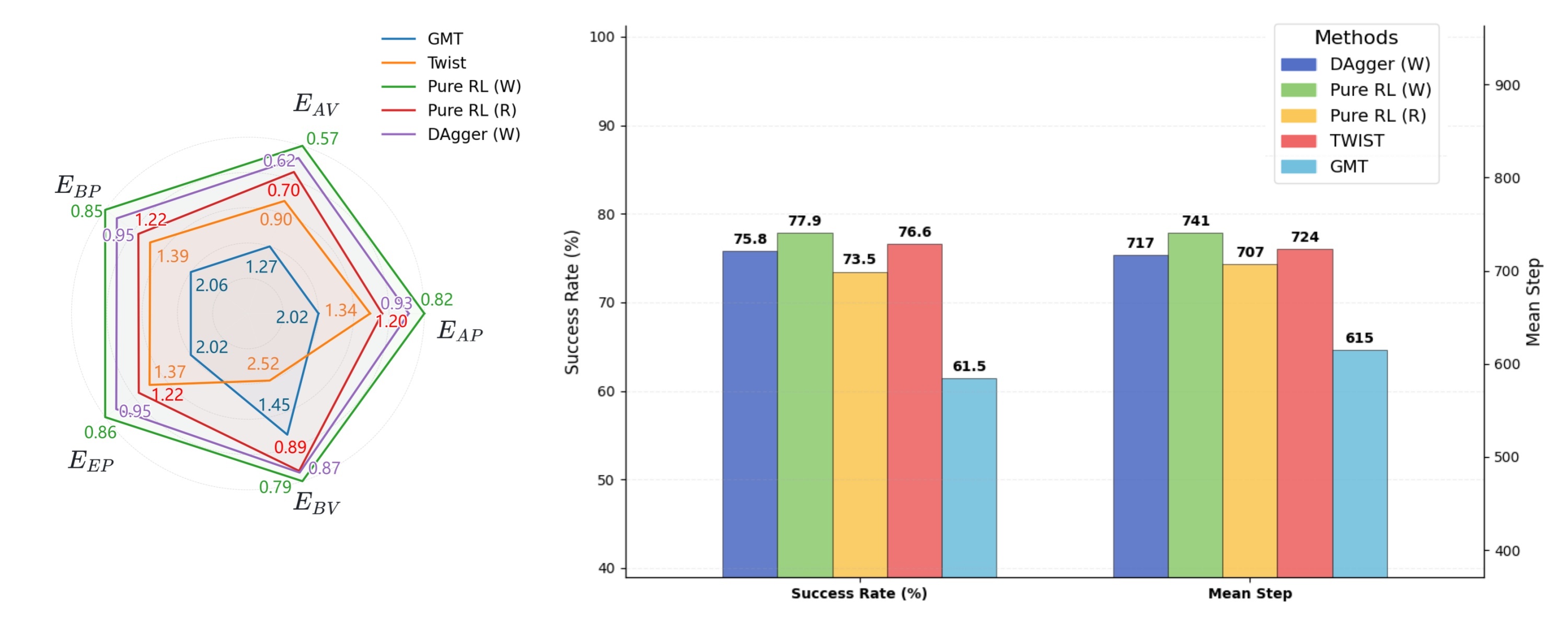}
        \caption{Quantitative Ablation of Reward Design and Training Paradigms Alongside Benchmarking with Prior Work.}
        \label{fig:compare_all}
    \end{subfigure}
    \caption{\textbf{Quantitative Comparison and Ablation Studies of General Motion Tracking.} The radar charts illustrate five core metrics characterizing tracking fidelity,
    while the bar charts depict robustness in terms of Success Rate and Average Steps per Episode. Fig \ref{fig:compare_multi_source} compares multi-source versus single-source data distributions. Fig \ref{fig:compare_all} evaluates our proposed variants (Pure RL + world frame reward, Pure RL + robot frame reward, and DAgger + world frame reward) against baselines GMT \cite{chen2025gmt} and TWIST \cite{ze2025twist}.}
    \label{fig:gmt_comparison}
\end{figure}


\begin{figure}[!t]
    \centering
    \includegraphics[width=1.0\linewidth]{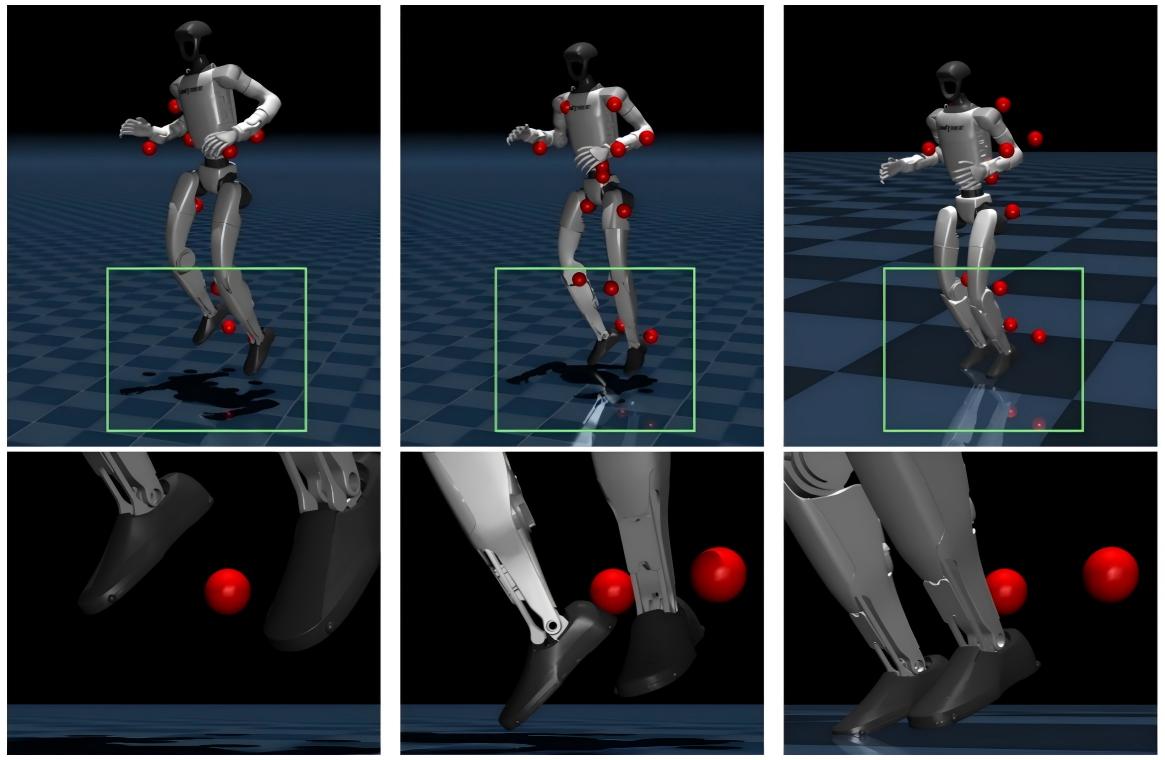}
    \caption{\textbf{Qualitative Comparison on High Dynamic Motion.} 
    From left to right are our model, TWIST, and GMT. Our model achieves substantial ground clearance at the reference apex, whereas baselines (i.e., TWIST and GMT) struggle to capture high-acceleration explosive movements.}
    \label{fig:jump_compare}
    \vspace{-1.7em}
\end{figure}

\section{Experiments}
\label{sec:experiments}

In the experimental section, we systematically investigate two core questions regarding robust general motion tracking and teleoperation to provide a mechanistic understanding.

\begin{enumerate}
    \item \textbf{How to train a robust general motion tracker for teleoperation-oriented tasks?}

    \item \textbf{Why do General Motion Trackers with saturated simulation metrics often underperform in real-world teleoperation, and how to efficiently adapt them using minimal teleoperation data without compromising general tracking robustness?}
\end{enumerate}

\subsection{Training Setting and Evaluation System}
\subsubsection{\textbf{Training Setting}}
Our policies are trained using the Isaac Lab~\cite{mittal2025isaaclab} framework on eight NVIDIA A100 GPUs over approximately 48 hours. To achieve efficient training, we leverage massive parallelism by instantiating 30,000 environments per GPU. The policies are parameterized by MLPs with hidden layers [1024,1024,512,256] and ELU activations \cite{clevert2015elu}.

\subsubsection{\textbf{Evaluation Metrics and Benchmark}}
We evaluate performance along two axes: tracking fidelity and robustness. Since mobile manipulation and teleoperation require faithful reproduction of the operator’s spatial displacement, we report all tracking metrics in the global (world) frame. To reflect long-duration usability, we additionally measure robustness via success rate and average steps per episode.
Concretely, our metrics include: (i) tracking fidelity---global anchor (torso) position error $E_{AP}$, anchor linear velocity error $E_{AV}$, body position error $E_{BP}$, body linear velocity error $E_{BV}$, and end-effector (EEF) position error $E_{EP}$; and (ii) robustness---success rate and average episode length (with maximum 500).

We benchmark in the MuJoCo physics engine~\cite{todorov2012mujoco} for sim-to-sim evaluation using the \textit{Online Videos} subset of Motion-X~\cite{lin2023motion} as an OOD test set. For a physically plausible benchmark, we curate \textit{Motion-X-Sub} by selecting sequences longer than 10\,s with maximum root height $\ge 0.6$\,m, filtering motions that require environmental support (e.g., sitting/swimming) or are excessively challenging. \textit{Motion-X-Sub} contains 633 sequences with an average duration of 19.04\,s, totaling $\sim$3.35\,hours of diverse motions.

\subsection{Design Principles for Robust General Motion Tracking}
To answer question 1), we conduct extensive comparative analyses across three dimensions: data sources, reward design, and training paradigms. Based on these comprehensive evaluations, we distill a set of key design principles for developing trackers that are both precise and robust.
\subsubsection{\textbf{Ablation on Data Source}}
To isolate the effect of data diversity, we construct three motion corpora with the same budget of 1M frames: \textit{1-Source} (AMASS~\cite{mahmood2019amass}), \textit{3-Sources} (AMASS + OMOMO~\cite{li2023omomo} + built in-house inertial MoCap data), and \textit{5-Sources} (further adding built in-house optical MoCap and GENMO-generated motions~\cite{li2025genmo}). All models are trained for the same iterations in identical configurations. 

As shown in Fig. \ref{fig:compare_multi_source}, increasing source diversity yields consistent gains across both tracking fidelity and robustness metrics.
Since evaluation is performed on an OOD test set, these improvements also indicate substantially better generalization from more heterogeneous motion distributions.

\subsubsection{\textbf{Ablation on Reward Design}}
BeyondMimic~\cite{liao2025beyondmimic} achieves strong single-motion tracking by emphasizing tracking objectives largely computed in the robot frame.
However, teleoperation-oriented mobile manipulation places stricter requirements on global tracking: purely local rewards can accumulate drift over long-horizon locomotion.
We therefore study the impact of augmenting ego-centric tracking with world-frame objectives that constrain global pose and motion.

Figure~\ref{fig:compare_all} shows that adding world-frame tracking substantially reduces long-horizon drift, reflected by lower global tracking errors, and improves stability by higher success rates and longer episode lengths.
We attribute the robustness gains to two coupled effects: (i) improved acquisition of dynamic locomotion patterns (e.g., walking, running and jumping), and (ii) an explicit incentive to maintain global trajectories, which discourages stumbles and falls that would otherwise disrupt target positions and velocities in world frame.

\subsubsection{\textbf{Ablation on Training Paradigm}}
We compare two common paradigms for motion tracking: pure RL and DAgger-style distillation~\cite{ross2011dagger}.
Our comparative evaluation demonstrates that pure RL consistently outperforms the distilled policy across several key dimensions.

As shown in Fig. \ref{fig:compare_all}, pure RL outperforms the distilled policy in tracking fidelity---precisely the metric of our primary concern. Furthermore, pure RL exhibits higher success rates and more average steps per episode, indicating enhanced robustness.
These results highlight a practical difficulty in distillation for motion tracking: transferring a teacher’s fine-grained global position/velocity tracking behavior into a deployable student without loss remains non-trivial. 
In contrast, when paired with sufficient data diversity and teleoperation-oriented rewards, pure RL directly optimizes long-horizon objectives and yields superior global tracking and stability, ultimately surpassing distillation-based alternatives in our setting.

\subsubsection{\textbf{Comparison with Representative Baseline}}
Finally, we compare our final tracker against prior representative methods, including GMT~\cite{chen2025gmt} and TWIST~\cite{ze2025twist}. 
As shown in Fig.~\ref{fig:compare_all}, our approach improves both tracking fidelity and robustness across all reported metrics, with particularly large gains in global position tracking. 
Qualitatively, the improvement is most apparent on high-dynamic segments: whereas prior methods often stumble or terminate during aggressive transitions, our policy maintains stable contacts and smooth state evolution. 
We include representative examples in Fig.~\ref{fig:jump_compare} to illustrate the resulting gains in tracking accuracy and motion quality.

\subsection{Bridging the Sim-to-Real Gap via Rapid Adaptation}
\label{sec:exp_adaptation}

We observe that while the general motion tracking policy achieves saturated metrics in simulation, its performance significantly degrades during real-world teleoperation. This discrepancy highlights a persistent sim-to-real gap. In this section, we analyze the sources of this gap and evaluate various strategies for rapid adaptation to various teleoperation scenarios to seek the answer to question 2).

\subsubsection{\textbf{Analysis of the Teleoperation Sim-to-Real Gap}}
\label{sec:analysis_gap}

Our evidence indicates that failures are driven less by limited motion diversity than by a teleoperation interface gap. Specifically:

\begin{itemize}
    \item \textbf{System latency.} End-to-end delays accumulate across sensing, transmission, processing, and control execution. According to measurements, VR streaming introduces approximately 400 ms latency, which creates a temporal mismatch between the operator’s intent and robot’s current state, leading to out-of-sync tracking and instability.
    \item \textbf{Noise and Jitter:} Real teleoperation streams exhibit packet loss, and sensor noise that are typically under-modeled in simulation. These artifacts introduce non-smooth perturbations in the reference trajectory, triggering motor safety limits or destabilize contact transitions.
\end{itemize}

A key empirical observation is that motion coverage is not the bottleneck. In simulation, the same tracker generalizes well to highly complex OOD motions; however, on robot we observe frequent stumbling even on fundamental gaits that are well represented in the training corpus. This contrast indicates that reliable deployment is primarily limited by teleoperation-specific interface effects (latency, noise, and hardware execution artifacts), rather than a lack of training diversity.

\begin{table}[htbp]
\centering
\caption{\textbf{Evaluation results on \textit{Motion-X-Sub}}. $\downarrow$ denotes lower is better, $\uparrow$ denotes higher is better.}
\label{tab:pico_motionx_results}
\resizebox{\columnwidth}{!}{
\begin{tabular}{lcccc}
\toprule
\textbf{Configuration} & \textbf{$E_{AP}$ (m)} $\downarrow$ & \textbf{$E_{BP}$ (m)} $\downarrow$ & \textbf{$E_{EP}$ (m)} $\downarrow$ & \textbf{Success Rate} $\uparrow$ \\
\midrule
Base Model       & 0.8242 & 0.8471 & 0.8578 & \textbf{77.88\%} \\
Fine-tune         & 2.7587 & 2.7387 & 2.6950 & 40.60\% \\
Continual        & 0.8571 & 0.8788 & 0.8870 & 78.36\% \\
Adapter (R)      & 0.8194 & 0.8412 & \textbf{0.8513} & 77.25\% \\
\textbf{Adapter (W)} & \textbf{0.8165} & \textbf{0.8406} & 0.8520 & 77.25\% \\
\bottomrule
\end{tabular}
}
\end{table}


\begin{table}[htbp]
\centering
\caption{\textbf{Evaluation results on VR dataset}.}
\label{tab:pico_dataset_results}
\resizebox{\columnwidth}{!}{
\begin{tabular}{lcccc}
\toprule
\textbf{Configuration} & \textbf{$E_{AP}$ (m)} $\downarrow$ & \textbf{$E_{BP}$ (m)} $\downarrow$ & \textbf{$E_{EP}$ (m)} $\downarrow$ & \textbf{Success Rate} $\uparrow$ \\
\midrule
Base Model       & 2.9352 & 2.9353 & 2.9363 & \textbf{100.00\%} \\
Fine-tune         & 1.4106 & 1.4043 & 1.3998 & 92.00\% \\
Adapter (R)      & 1.5278 & 1.5370 & 1.5436 & 96.00\% \\
Adapter (W)      & \textbf{1.1940} & \textbf{1.1961} & \textbf{1.2002} & \textbf{100.00\%} \\
Continual        & 1.7248 & 1.7272 & 1.7294 & \textbf{100.00\%} \\
\bottomrule
\end{tabular}
}
\end{table}


\begin{table}[htbp]
\centering
\caption{\textbf{Data scaling analysis on VR dataset}.}
\label{tab:data_scaling_results}
\resizebox{\columnwidth}{!}{
\begin{tabular}{lcccc}
\toprule
\textbf{Configuration} & \textbf{$E_{AP}$ (m)} $\downarrow$ & \textbf{$E_{BP}$ (m)} $\downarrow$ & \textbf{$E_{EP}$ (m)} $\downarrow$ & \textbf{Success Rate} $\uparrow$ \\
\midrule
Adapter (3 min)   & 2.8254 & 2.8257 & 2.8269 & \textbf{100.00\%} \\
Adapter (15 min)  & 1.5686 & 1.5674 & 1.5651 & 96.00\% \\
\textbf{Adapter (30 min)} & \textbf{1.1940} & \textbf{1.1961} & \textbf{1.2002} & \textbf{100.00\%} \\
\midrule
FLD Augmentation  & 2.9298 & 2.9298 & 2.9307 & \textbf{100.00\%} \\
\bottomrule
\end{tabular}
}
\end{table}

\subsubsection{\textbf{Exploration of Strategies for Rapid Adaptation}}
To bridge the interface gap, we collected 30 minutes of VR teleoperation data and explored three adaptation strategies based on the same pre-trained general motion tracker (GMT):
\begin{itemize}
    \item \textbf{Fine-tuning}: Continue optimizing $\pi_{GMT}$ exclusively on the teleoperation dataset.
    \item \textbf{Continual Learning}: Mix the teleoperation data with the original motion corpus and continue training.
    \item \textbf{Residual Adapter}: Sec.~\ref{sec:method}, Reward in world frame: Adaptor (W), and reward in robot frmae: Adaptor (R).
\end{itemize}

\paragraph{Preserving Generalist Capabilities} We first evaluate whether adaptation compromises general tracking on \textit{Motion-X-Sub}. As reported in Table~\ref{tab:pico_motionx_results}, Residual Adapter yields the strongest overall tracking and robustness metrics, indicating that interface-specific corrections can be injected without degrading the general tracker’s broad coverage.

\paragraph{Real-Robot Teleoperation Performance} We further evaluated these strategies on a 10-minute held-out real-robot teleoperation test set. As shown in Table~\ref{tab:pico_dataset_results}, the Residual Adapter consistently outperforms all baselines. Fine-tuning yields only marginal position-tracking gains but substantially reduces robustness, suggesting overfitting to the small adaptation set. Continual Learning provides negligible improvement, likely because the limited real data is dominated by the large general-motion corpus (gradient dilution). Overall, the Residual Adapter enables rapid, data-efficient adaptation while preserving the base policy’s general performance.


\subsubsection{\textbf{Further Exploration of the Residual Adapter}}
\paragraph{Data Scaling} To investigate the data requirements for residual adaptation, we vary the adaptation dataset sizes: 3, 15, and 30 minutes. Experimental results indicate that 3 minutes of data is insufficient to produce meaningful improvement. Gains emerge at 15 minutes and further increase at 30 minutes, suggesting that a modest amount of high-quality teleoperation data can efficiently bridge the sim-to-real gap.

\paragraph{Representative-based data augmentation}

As an alternative to mitigate long-horizon locomotion sparsity, we explored structured motion synthesis with Fourier Latent Dynamics (FLD)~\cite{li2024fld}. We train FLD on 3 minutes of periodic teleoperated locomotion and synthesize $\sim$10 hours of trajectories via latent-space sampling, then train trackers with and without FLD augmentation under matched settings and distill both into the residual policy for controlled comparison. FLD improves interpolation within the periodic manifold (e.g., unseen gait frequencies and step lengths), but yields only marginal gains on the held-out teleoperation test: failures on complex OOD motions remain. Overall, in our setting, interface-level adaptation is more effective for deployment robustness than expanding periodic locomotion data.
\section{Conclusions and Limitations}
\label{sec:discussion}

\paragraph{Conclusions}
We presented MOSAIC, a humanoid motion tracking and teleoperation system that combines a general motion tracker with a data-efficient residual interface adaptor. Our study yields three key insights: (\textbf{A}) after simulation tracking metrics saturate, real-robot failures are driven mainly by interface and dynamics gaps (e.g., latency, estimation bias, and teleoperation noise), rather than insufficient motion diversity; (\textbf{B}) merging a small amount of teleoperation data into large-scale training can cause forgetting or gradient dilution, whereas residual adaptation injects new capability as incremental correction on a frozen backbone and is the most stable and sample-efficient in our setting; (\textbf{C}) periodic-motion augmentation (e.g., FLD) improves motion coverage but does not address interface-induced system errors, yielding smaller deployment gains than direct adaptation. Overall, separating general tracking from teleoperation-specific correction enables minute-scale stable teleoperation while preserving broad motion generality and supporting rapid adaptation to new interfaces with minimal additional data.

\paragraph{Limitations}
MOSAIC does not eliminate dependence on reliable low-latency sensing and state estimation, and residual adaptation primarily targets interface shifts rather than all sources of sim-to-real mismatch.

\bibliographystyle{plainnat}
\bibliography{references}

\newpage

\appendix
\setcounter{figure}{0}
\setcounter{table}{0}
\renewcommand{\thefigure}{S\arabic{figure}}
\renewcommand{\thetable}{S\arabic{table}}
\section*{Dataset Details}
\label{sec:data}
\subsection{Dataset Sources}
\begin{figure*}[!t]
    \centering
    \includegraphics[width=\linewidth]{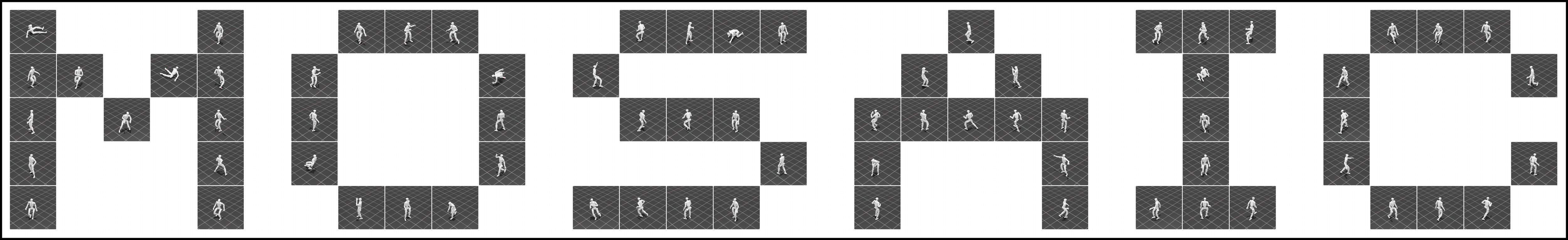}
    \caption{\textbf{Randomly Sampled Motions From the Motion Dataset.} Each tile visualizes a motion frame extracted at random time indices across different motions, illustrating the diversity of the motion dataset.}
    \label{fig:mosaic}
\end{figure*}

\paragraph{Multi-source motion corpus.}
We train and evaluate MOSAIC using a heterogeneous motion corpus that combines large-scale public datasets, in-house motion capture, and curated synthetic motions. Fig~\ref{fig:mosaic} shows some randomly sampled motions. 
\textbf{Public corpora} include AMASS~\cite{mahmood2019amass} and OMOMO~\cite{li2023omomo}, totaling 51 hours of diverse human motions. 
In-house captures consist of (i) \textbf{optical MoCap} recordings: we collected 6 hours in total and retained 3.1 hours after strict quality control (e.g., removing marker swaps, severe occlusion, and inconsistent contacts), and (ii) \textbf{inertial MoCap} recordings totaling 7 hours. 
\textbf{Generated motions} are obtained via GENMO~\cite{li2025genmo}: we synthesized roughly 4 hours of motions and manually curated 2.2 hours of high-quality clips for training. 
Finally, for \textbf{interface adaptation} and evaluation, we collected teleoperation demonstrations from two distinct interfaces, each providing approximately 0.5 hours of data.

\paragraph{Different interface between training and deployment.}
Importantly, the 7-hour inertial MoCap dataset used for training the general motion tracker is captured with a different commercial suit than the device used in our final teleoperation tests (training: inertial MoCap suit from IO-AI~\cite{io_web}.Tech; testing: inertial MoCap suit from Noitom.Tech~\cite{noitom_axis}). 
This intentional device mismatch induces realistic interface-specific latency, noise, and kinematic bias, and motivates our residual adaptation module, which injects interface corrections without overwriting the generalist tracking capability learned from the multi-source motion bank.

\subsection{Data Processing}
\subsubsection{Built-in House Data}
\paragraph{\textbf{Optical Motion Capture Pipeline.}} 
We developed a comprehensive pipeline to transform raw optical motion capture data into standardized formats for robot learning. The workflow is categorized into four main stages:

\begin{enumerate}
    \item \textbf{Data Acquisition and Pre-processing:} 
    We utilize the Vicon Nexus \cite{vicon_nexus} system to capture human motions with a 39-marker setup at a sampling rate of 100 Hz.. The specific marker placement is illustrated in Figure~\ref{fig:marker_layout}. Each captured sequence is paired with a corresponding natural language description, same as the prompts using for motion generation. The raw trajectories undergo a rigorous pre-processing phase, including automated segment labeling followed by manual correction, gap filling for occluded markers, and low-pass filtering to eliminate high-frequency noise. The refined data are exported in the \texttt{.c3d} format.

    \item \textbf{Human Motion Reconstruction and Quality Control:} 
    To convert marker clouds into parametric human poses, we employ Mosh++ to generate \texttt{.pkl} files compatible with the AMASS framework. To ensure data fidelity, we perform a manual quality inspection using a MuJoCo-based replay script \cite{todorov2012mujoco}. Any sequences exhibiting physical inconsistencies, such as joint jitters or unnatural discontinuities, are strictly excluded. The validated data are subsequently converted into \texttt{.npz} format.

    \item \textbf{Motion Retargeting:} 
    The validated human motions are retargeted to the Unitree G1 humanoid robot~\cite{unitree_g1_web} using the GMR \cite{ze2025gmr} framework. This process maps human kinematics onto the G1's degrees of freedom (DoF), producing robot-specific pose sequences stored in \texttt{.pkl} format.

    \item \textbf{Final Formatting and Standardization:}
    In the final stage, the retargeted motion sequences in \texttt{.pkl} format are converted to \texttt{.csv} and subsequently transformed into the final \texttt{.npz} format using the BeyondMimic \cite{liao2025beyondmimic} processing suite to meet the requirements of the training framework.
\end{enumerate}

\begin{figure}[!t]
    \centering
    \includegraphics[width=1.0\linewidth]{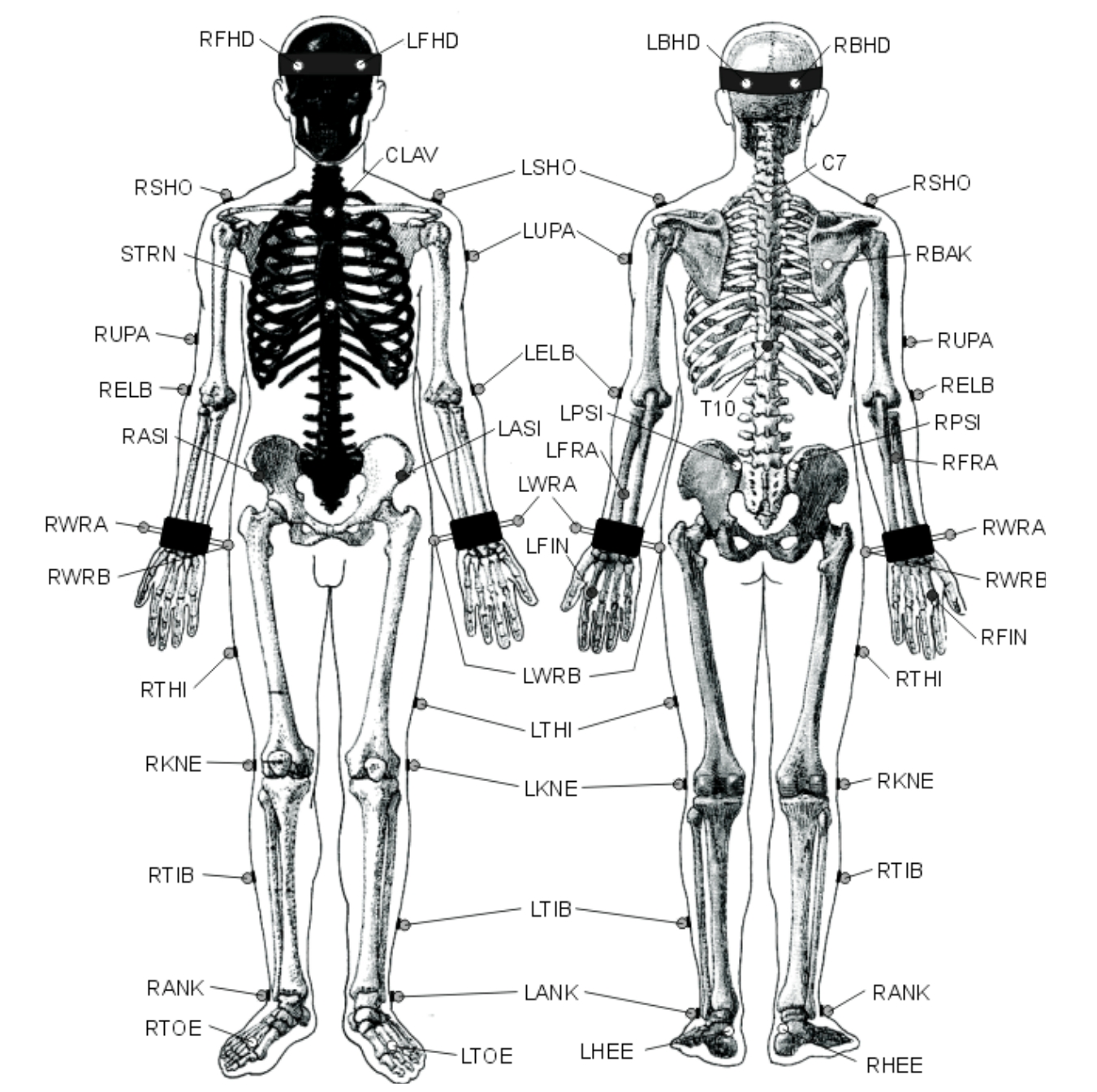}
    \caption{\textbf{Schematic of the Full-body Optical Marker Configuration.} To ensure high precision and motion fidelity during data acquisition, we adopt a comprehensive setup consisting of 39 reflective markers.}
    \label{fig:marker_layout}
\end{figure}

\paragraph{\textbf{Inertial Motion Capture Pipeline.}}
In addition to optical methods, we employ a high-fidelity inertial motion capture pipeline to expand our dataset. The workflow is structured as follows:

\begin{enumerate}
    \item \textbf{Data Acquisition and Hardware Configuration:}
    We utilize the IO-AI~\cite{io_web} full-body inertial motion capture system, which comprises 15 high-precision IMU sensors. To capture fine-grained manipulations and full-body coordination, the sensors are distributed with 9 on the upper body and 6 on the lower body, operating at a sampling frequency of 120 Hz. Raw data are initially captured onto local portable storage and subsequently uploaded asynchronously. This protocol effectively eliminates packet loss and temporal drift typically associated with wireless transmission, ensuring the integrity of long-duration motion sequences.
    
    \item \textbf{Standardized Data Management:} 
    Data collection is managed through a standardized platform where task specifications and metadata are pre-defined. During acquisition, relevant task information is automatically bound to the motion sequences. This automated metadata integration reduces manual labeling requirements while enhancing data consistency and traceability across large-scale collection sessions.
    
    \item \textbf{Downstream Processing and Standardization:} 
    The acquired AMASS-compatible sequences are processed following the same downstream pipeline as the optical data. This involves retargeting human kinematics to the Unitree G1 robot via the GMR \cite{ze2025gmr} framework, followed by final standardization into the \texttt{.npz} format using the BeyondMimic \cite{liao2025beyondmimic} suite.
\end{enumerate}

\begin{figure}[!t]
    \centering
    \includegraphics[width=1.0\linewidth]{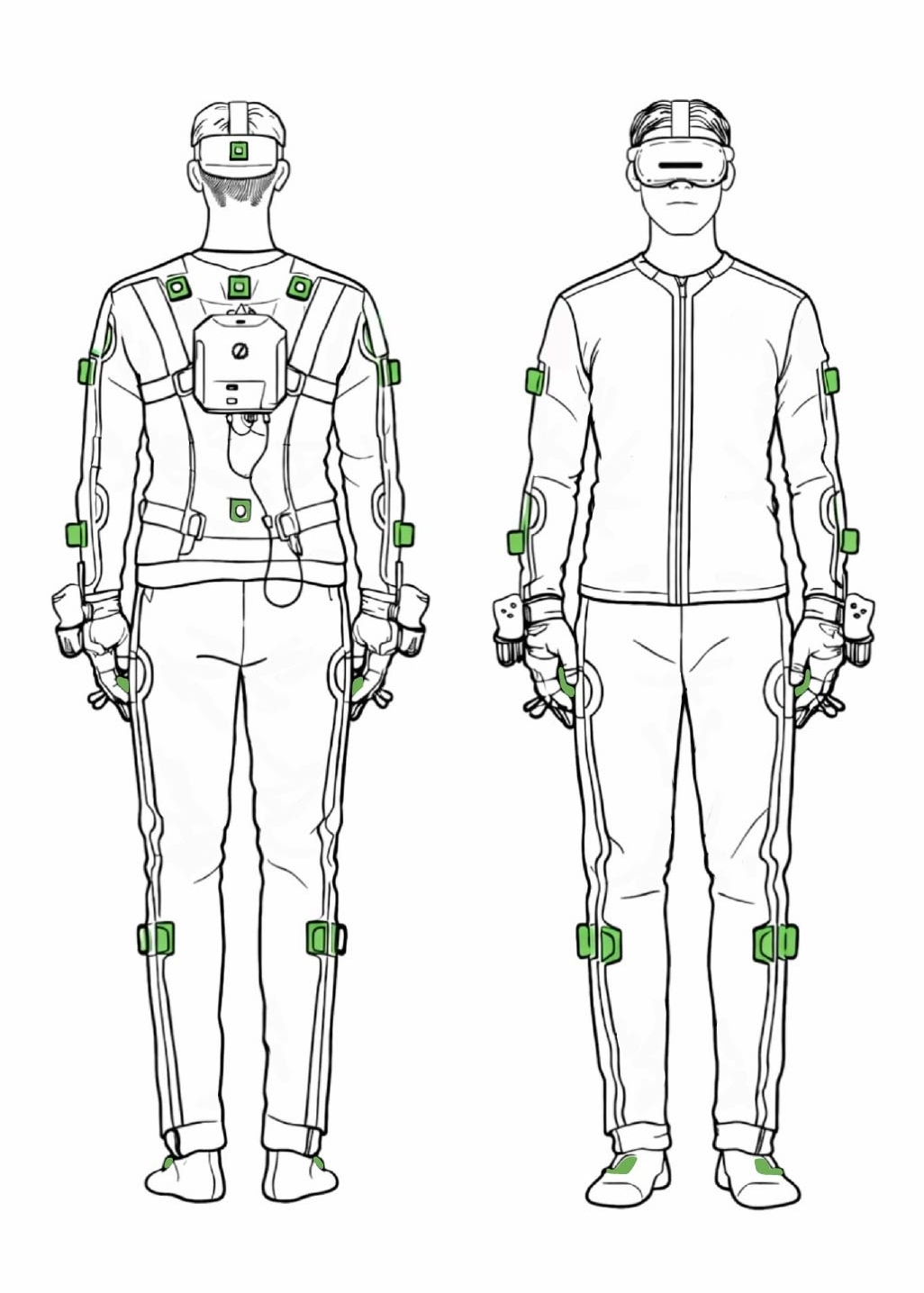}
    \caption{\textbf{Schematic of the Full-body Inertial Sensors Configuration.} To ensure high precision and motion fidelity during data acquisition, we utilize a comprehensive setup consisting of 15 IMU sensors, which operate in coordination with a VR headset and handheld controllers.}
    \label{fig:marker_layout}
\end{figure}

\subsubsection{Public Motions}
Our public-motion corpus is built from the AMASS~\cite{mahmood2019amass} and OMOMO~\cite{li2023omomo} datasets, which we download directly from their official project pages. We treat these datasets identically to our in-house captures: after acquisition, all sequences are retargeted to our robot kinematic model and processed through the same standardization pipeline as the built-in house data. Concretely, we unify coordinate conventions and units, align initial poses to a consistent canonical configuration (e.g., root orientation and heading), and apply the same representation/format conversion required by our training interface (e.g., robot-space joint trajectories and body kinematics at a fixed rate). As a result, public and in-house motions share a single, consistent on-robot format and can be mixed seamlessly during training and evaluation.
\subsubsection{Generated Motions}
To augment motion diversity in a controllable manner, we also construct a synthetic motion set using a text-to-motion pipeline. First, we use an LLM (ChatGPT) to generate text prompts spanning seven categories: (i) static poses, (ii) sports/exercises, (iii) running, (iv) daily activities, (v) martial arts, (vi) dancing, and (vii) walking. 
All prompts explicitly constrain motions to be planar (flat-ground behaviors) and require a duration of at least 10\,s. 
Second, we feed these prompts to GENMO~\cite{li2025genmo}'s text-to-motion model to synthesize AMASS-format motion sequences. 
Third, we retarget the synthesized motions into the robot-space representation using GMR, and apply the same coordinate normalization and initialization alignment used for real MoCap data (e.g., consistent root frame conventions and initial pose alignment). 
Finally, we manually curate the generated set to remove low-quality sequences, including motions with severe self-collisions, discontinuities, or physically implausible transitions. 
After curation, we retain approximately 800 high-quality motion clips for training.

\subsection{Motion Representation and Sampling Strategy}
\paragraph{Motion Representation.}
Each motion clip is stored as an \texttt{.npz} file containing a fixed set of tensors at a common frame rate \texttt{fps} (50\,Hz in our training). The motion loader reads the following fields: joint positions/velocities and body kinematics in world frame for a selected set of bodies. The representation is already in the robot joint space, and thus directly matches the policy’s tracking targets and the simulator’s state.

To ensure fast training throughput, we preload all motion files into contiguous tensors. For motion $m$, we record its length $L_m$ and maintain a cumulative offset $O_m$ so that any per-motion frame index can be mapped to a global index:
\begin{equation}
g(m,t)=O_m+\min(t, L_m-1).
\end{equation}
At runtime, batched queries are served by indexed gathers into these contiguous tensors, eliminating per-step file I/O and enabling high-throughput sampling from large motion corpora.

We additionally support multi-GPU sharding at load time, which deterministically selects a disjoint subset of motion files per rank, reducing duplicates and memory overhead in distributed training.

\paragraph{Two-Level Adaptive Resampling.}
Our resampling operates at two levels:
\begin{enumerate}[leftmargin=1.2em]
    \item \textbf{Motion-level assignment:} periodically map each environment to a motion index (which clip is currently tracked).
    \item \textbf{Within-motion time sampling:} when an environment resets (or reaches the end of a clip), sample a time step inside that motion using failure-aware bin probabilities.
\end{enumerate}
Both mechanisms are implemented inside the command class, which serves the tracking command by gathering a fixed-horizon future window.

\subsubsection{Fixed-horizon window sampling}
At every step, the policy receives a fixed-horizon reference window of length $H$ frames (in our setting, $H=1$). Given the environment’s current motion index $m_e$ and time step $t_e$, we gather frames $\{t_e, t_e+1,\dots,t_e+H-1\}$ (clamped at $L_m-1$) and return the flattened sequence
\begin{equation}
\mathbf{y}_{e} = [q^{ref}_{t:t+H-1}, \dot q^{ref}_{t:t+H-1}] \in \mathbb{R}^{H\times D},
\end{equation}
which keeps the policy interface consistent across heterogeneous datasets with highly uneven clip lengths.

\subsubsection{Motion-level sampling: difficulty $\times$ novelty $\times$ uniform}
When remapping environments to motions, we sample motion indices from a probability vector. The distribution is a convex mixture of three terms.

\textbf{Difficulty from motion-level failure rate.}
We maintain per-motion counters: motion\_sample\_counts $S_m$ (times the motion is resampled), motion\_fail\_counts $F_m$ (times episodes terminated while tracking that motion), and compute the mean failure rate $r_m$:
\begin{equation}
r_m=\frac{F_m}{S_m+\epsilon}.
\end{equation}
To reduce the effect of extreme early failures, we cap the rates using cap\_beta $\beta$ and the mean failure rate $r_m$:
\begin{equation}
\tilde r_m=\min\!\big(r_m, \beta\,\mathbb{E}[r]\big),
\end{equation}
and normalize to obtain failure probability $p_{\mathrm{fail}}(m)\propto \tilde r_m$.

\textbf{Novelty from assignment count.}
To encourage coverage, we add a novelty term that favors less-assigned motions:
\begin{equation}
p_{\mathrm{novel}}(m)\propto \frac{1}{\sqrt{A_m+1}},
\end{equation}
where $A_m$ is the count of assigned motions.

\textbf{Uniform coverage.}
We always include a uniform baseline $p_{\mathrm{uni}}(m)=1/M$, where $M$ is the number of all motions.

The final distribution is:
\begin{equation}
\begin{split}
    p(m)=w_f\,p_{\mathrm{fail}}(m)+w_n\,p_{\mathrm{novel}}(m)+w_u\,p_{\mathrm{uni}}(m), \\
    w_u=1-w_f-w_n.
\end{split}
\end{equation}
The implementation ramps $w_f,w_n$ from 0 to target values (the weight for failure, the weight for novelty) using a warmup + ramp schedule. This mitigates cold-start bias where early failures are ubiquitous and difficulty estimates are uninformative.

\textbf{Active-motion pool.}
When the motion database is much larger than the number of environments, we optionally sample an active pool of $K$ motions without replacement and assign them to environments approximately uniformly. This improves memory locality and ensures each selected motion receives sufficient training signal before the next remap.

\textbf{Remap cadence.}
We periodically reassign motions every interval of resampling motions so that, over training, all motions get repeated chances to be tracked.

\paragraph{Note on dataset-level weighting.}
In our setting, motion-level sampling operates on a single folder of motions. Dataset-level weighting can be realized by (i) constructing the motion folder with desired per-source proportions, or (ii) extending the probability of motion sampling with per-motion weights (e.g., a scalar multiplier based on source ID). The remainder of the pipeline (warmup/ramp, novelty, and within-motion bin sampling) applies unchanged.

\subsubsection{Within-motion failure-aware bin resampling (EMA + kernel smoothing)}
Even with motion-level balancing, failures often concentrate in specific segments (e.g., sharp turns, contact transitions). We implement within-motion adaptive sampling following BeyondMimic~\cite{liao2025beyondmimic}.

\section*{Training Details}
\label{sec:train}

\subsection{Observations}
The observation space is divided into proprioceptive data for the actor and privileged information for the critic, as illustrated in Table ~\ref{tab:obs_noise}. To enhance robustness against sensor inaccuracies, the actor processes a 5-step history of proprioceptive signals corrupted by Gaussian noise. In contrast, the critic utilizes the noise-free, ground-truth observations of the current frame to provide a more accurate value estimation during training.

\begin{table}
    \centering
    \small
    \renewcommand{\arraystretch}{1.2}
    \begin{tabularx}{\columnwidth}{l X c}
    \toprule
    \textbf{Category} & \textbf{Observation Term} & \textbf{Noise} \\
    \midrule
    \multirow{7}{*}{Prop.} 
    & reference joint position                 & --              \\
    & reference joint velocity                 & --              \\
    & anchor orientation error                 & $\mathcal{U}(-0.05,\,0.05)$ \\
    & robot base angular velocity              & $\mathcal{U}(-0.2,\,0.2)$   \\
    & robot joint position                     & $\mathcal{U}(-0.01,\,0.01)$ \\
    & robot joint velocity                     & $\mathcal{U}(-0.5,\,0.5)$   \\
    & last action                              & -- \\
    \midrule
    \multirow{5}{*}{Priv.} 
    & anchor position error                    & -- \\
    & body position (robot frame)              & -- \\
    & body orientation (robot frame)           & -- \\
    & robot base linear velocity               & -- \\
    & reference base linear velocity           & -- \\
    \bottomrule
    \end{tabularx}
    \caption{\textbf{Observation Terms and Noises Used During Policy Training.} 
    The actor uses the \textit{Prop} observation stack with a 5-step history, while the critic uses the full set of observations including \textit{Prop} and \textit{Priv} terms at the current step. Here, the robot's \textit{torso link} is defined as the \textit{anchor} for all coordinate transformations. Accordingly, the \textit{anchor position error} and \textit{anchor orientation error} are represented as the target reference orientation expressed in the robot's local frame.}
    \label{tab:obs_noise}
\end{table}

\subsection{Rewards}
The reward function consists of tracking rewards (Global and Teleop style) and penalties for regularization, as shown in Table ~\ref{tab:reward_spec_final}. Tracking rewards are calculated using an exponential kernel: $r = w \cdot \exp(-\|e\|^2 / \sigma^2)$.

\begin{table*}[t]
    \centering
    \caption{\textbf{Detailed Reward Function Specifications.} All tracking rewards utilize the kernel $r = \exp(-\text{error} / std^2)$. The total reward is the weighted sum of tracking accuracy and regularization penalties.}
    \label{tab:reward_spec_final}
    \small
    \renewcommand{\arraystretch}{1.3}
    
    \begin{tabularx}{\textwidth}{l >{\raggedright\arraybackslash}X c l}
    \toprule
    \textbf{Reward Term} & \textbf{Equation} & \textbf{Weight} & \textbf{Parameters} \\
    \midrule
    \textit{Tracking rewards} & & & \\
    Motion global anchor position & $r_{\text{pos}}^{\text{anc}} = \exp( - \|\mathbf{p}_{a}^p - \mathbf{p}_{a}^g\|_2^2 / std^2 )$ & +0.5 & std=0.3 \\
    Motion global anchor orientation & $r_{\text{ori}}^{\text{anc}} = \exp( - d_{quat}(\mathbf{q}_{a}^p, \mathbf{q}_{a}^g)^2 / std^2 )$ & +0.5 & std=0.4 \\
    Motion body position & $r_{\text{pos}}^{\text{body,rel}} = \exp( - \text{mean}_{b \in \mathcal{B}} \|\mathbf{p}_{b}^{\text{rel},p} - \mathbf{p}_{b}^{\text{rel},g}\|_2^2 / std^2 )$ & +1.0 & std=0.3 \\
    Motion body orientation & $r_{\text{ori}}^{\text{body,rel}} = \exp( - \text{mean}_{b \in \mathcal{B}} d_{quat}(\mathbf{q}_{b}^{\text{rel},p}, \mathbf{q}_{b}^{\text{rel},g})^2 / std^2 )$ & +1.0 & std=0.4 \\
    Motion body linear velocity & $r_{\text{lin}}^{\text{body}} = \exp( - \text{mean}_{b \in \mathcal{B}} \|\mathbf{v}_{b}^p - \mathbf{v}_{b}^g\|_2^2 / std^2 )$ & +1.5 & std=1.0 \\
    Motion body angular velocity & $r_{\text{ang}}^{\text{body}} = \exp( - \text{mean}_{b \in \mathcal{B}} \|\boldsymbol{\omega}_{b}^p - \boldsymbol{\omega}_{b}^g\|_2^2 / std^2 )$ & +1.5 & std=3.14 \\
    Motion anchor linear velocity & $r_{\text{vel}}^{\text{anc}} = \exp( - \|\mathbf{v}_{a}^p - \mathbf{v}_{a}^g\|_2^2 / std^2 )$ & +1.0 & std=1.0 \\
    \midrule
    \textit{Teleoperation rewards} & & & \\
    Motion global body position & $r_{\text{pos}}^{\text{ext}} = w_u \exp( - \text{err}_{u} / std^2 ) + w_l \exp( - \text{err}_{l} / std^2 )$ & +1.0 & std=0.5 \\
    Motion global VR & $r_{\text{vr}} = \exp( - \text{mean}_{k \in \mathcal{K}} \|\mathbf{p}_{k}^p - \mathbf{p}_{k}^g\|_2^2 / std^2 ), \mathcal{K}=\{\text{head, hands}\}$ & +0.5 & std=0.5 \\
    Motion global feet position & $r_{\text{feet}} = \exp( - \text{mean}_{f \in \mathcal{F}} \|\mathbf{p}_{f}^p - \mathbf{p}_{f}^g\|_2^2 / std^2 ), \mathcal{F}=\{\text{ankles}\}$ & +1.0 & std=0.5 \\
    Motion global body orientation & $r_{\text{rot}}^{\text{ext}} = \exp( - \text{mean}_{b \in \mathcal{B}} d_{quat}(\mathbf{q}_{b}^p, \mathbf{q}_{b}^g)^2 / std^2 )$ & +0.5 & std=0.5 \\
    Motion global body angular velocity & $r_{\text{ang}}^{\text{ext}} = \exp( - \text{mean}_{b \in \mathcal{B}} \|\boldsymbol{\omega}_{b}^p - \boldsymbol{\omega}_{b}^g\|_2^2 / std^2 )$ & +0.5 & std=3.14 \\
    Motion global body linear velocity & $r_{\text{lin}}^{\text{ext}} = \exp( - \text{mean}_{b \in \mathcal{B}} \|\mathbf{v}_{b}^p - \mathbf{v}_{b}^g\|_2^2 / std^2 )$ & +0.5 & std=1.0 \\
    \midrule
    \textit{Penalty terms} & & & \\
    Undesired contacts & $r_{\text{contact}} = \sum_{c \notin \{\text{feet, hands}\}} \mathbb{1}[\|\mathbf{F}_c\| > \text{threshold}]$ & -0.05 & threshold=1.0N \\
    Action rate L2 & $r_{\text{act}} = \|\mathbf{a}_t - \mathbf{a}_{t-1}\|_2^2$ & -0.1 & -- \\
    Joint limit & $r_{\text{jlim}} = \sum_{j} \mathbb{1}[\mathbf{q}_{t,j} \notin [\mathbf{q}_{j}^{\text{min}}, \mathbf{q}_{j}^{\text{max}}]]$ & -10.0 & -- \\
    Joint acceleration & $r_{\text{acc}} = \|\ddot{\mathbf{q}}_t\|_2^2$ & -2.5e-7 & -- \\
    Joint torque & $r_{\text{torque}} = \|\boldsymbol{\tau}_t\|_2^2$ & -1e-5 & -- \\
    \bottomrule
    \end{tabularx}
    \vspace{0.3em}
    {\footnotesize\textit{Note:} $w_u$ and $w_l$ denote the weights for the upper-body and lower-body components, respectively; $\mathrm{err}_u$ and $\mathrm{err}_l$ are the corresponding tracking errors.}
\end{table*}

\subsection{Terminations}
Terminations ensure the safety and efficiency of the training process by resetting episodes that violate pre-defined constraints. Details are illustrated in Table ~\ref{tab:terminations}.

\begin{table}[t]
    \centering
    \small
    \renewcommand{\arraystretch}{1.2}
    \begin{tabularx}{\columnwidth}{l X}
    \toprule
    \textbf{Termination Term} & \textbf{Threshold / Condition} \\
    \midrule
    Motion End                & Completion of the target motion sequence \\
    Time Out                  & Reaching maximum episode length \\
    Anchor Pos. Error         & Vertical error $|z_{anchor} - z_{ref}| > 0.25$ m \\
    Anchor Ori. Error         & Orientation deviation $> 0.8$ rad \\
    EE Body Pos. Error        & Vertical error $|z_{ee} - z_{ref}| > 0.25$ m for wrists/ankles \\
    \bottomrule
    \end{tabularx}
    \caption{\textbf{Termination Conditions During Policy Training.}}
    \label{tab:terminations}
\end{table}

\subsection{Domain Randomization}
To facilitate sim-to-real transfer, we randomize physical properties at the start of each episode and apply random pushes at random intervals of 1–3 seconds during simulation. Details are shown in Table ~\ref{tab:domain_rand}.

\begin{table}[t]
    \centering
    \small
    \renewcommand{\arraystretch}{1.2}
    \begin{tabularx}{\columnwidth}{l X l}
    \toprule
    \textbf{Parameter} & \textbf{Randomization Range} & \textbf{Mode} \\
    \midrule
    Static Friction    & $[0.3, 1.6]$                 & Startup \\
    Dynamic Friction   & $[0.3, 1.2]$                 & Startup \\
    Restitution        & $[0.0, 0.5]$                 & Startup \\
    Joint Default Pos  & $\text{Base} \pm 0.01$ rad   & Startup \\
    Base CoM Offset    & $x \in \pm 0.025, y/z \in \pm 0.05$ m & Startup \\
    Robot Push         & Sampling velocity range      & Interval (1-3s) \\
    \bottomrule
    \end{tabularx}
    \caption{\textbf{Domain Randomization Configuration.}}
    \label{tab:domain_rand}
\end{table}

\subsection{PPO Parameters}
We train all RL policies with PPO using a shared set of hyperparameters and an MLP actor--critic within the framework RSL\_RL~\cite{schwarke2025rslrl}.
Details are summarized in Table~\ref{tab:ppo_params}.

\begin{table}[t]
    \centering
    \small
    \renewcommand{\arraystretch}{1.2}
    \begin{tabularx}{\columnwidth}{l X}
    \toprule
    \textbf{PPO / Policy Parameter} & \textbf{Value} \\
    \midrule
    Rollout length (per env, per update) & \(24\) \\
    Learning epochs & \(5\) \\
    Mini-batches per epoch & \(4\) \\
    Learning rate & \(10^{-3}\) \\
    LR schedule & adaptive \\
    desired KL & \(0.01\) \\
    Discount factor \(\gamma\) & \(0.99\) \\
    GAE parameter \(\lambda\) & \(0.95\) \\
    PPO clip ratio \(\epsilon\) & \(0.2\) \\
    Value loss coefficient & \(1.0\) \\
    Entropy coefficient & \(0.005\) \\
    Max gradient norm & \(1.0\) \\
    Actor MLP hidden dims & \([1024,1024,512,256]\) \\
    Critic MLP hidden dims & \([1024,1024,512,256]\) \\
    Activation & ELU \\
    Initial action std & \(1.0\) \\
    \bottomrule
    \end{tabularx}
    \caption{\textbf{PPO Hyperparameters Used for Policy Training.}}
    \label{tab:ppo_params}
\end{table}

\subsection{Residual Adaptor}
\paragraph{Residual module architecture.}
The residual adaptor is a lightweight MLP that maps deployable actor observations to an action-space correction.
It consists of three hidden layers with widths \([512,256,128]\) and ELU activations, followed by a linear output layer producing a residual action \(\Delta\mathbf{a}_t\).
The adaptor shares the same action space as the backbone and is added to the frozen backbone action at every timestep.

\begin{table}[t]
    \centering
    \small
    \renewcommand{\arraystretch}{1.2}
    \begin{tabularx}{\columnwidth}{l X}
    \toprule
    \textbf{Item} & \textbf{Specification} \\
    \midrule
    MLP Hidden dims &\([512,256,128]\) \\
    Activation & ELU \\
    Output layer & linear, small-gain Xavier init (gain \(0.01\)) with zero bias \\
    \bottomrule
    \end{tabularx}
    \caption{\textbf{Residual Adaptor Architecture.}}
    \label{tab:residual_arch}
\end{table}

\section*{Deployment Details}
\label{sec:deploy}

\subsection{Real Robot Setup}
\paragraph{Robot platform and low-level actuation.}
All real-robot experiments in this paper use the Unitree G1 humanoid~\cite{unitree_g1_web} with 29 actuated DoFs. 
The high-level policy runs at 50\,Hz and outputs full-body joint position targets. 
Targets are executed by a standard joint-space PD impedance controller implemented via Isaac Lab implicit actuators, with joint-wise torque (effort) and velocity limits for safety and hardware fidelity, following the actuation setting of BeyondMimic~\cite{liao2025beyondmimic}: We parameterize PD gains using a second-order system interpretation, specifying a desired closed-loop natural frequency ($\omega_n$) and damping ratio ($\zeta$) . 
For each actuator group, stiffness and damping are computed from the joint armature (effective inertia) $J$ as
$
K_p = J \omega_n^2,\;
K_d = 2 \zeta J \omega_n,
$
where we set $\omega_n = 2\pi\cdot 10\,\mathrm{Hz}$ and $\zeta = 2.0$. 
This yields consistent impedance behavior across joints with different effective inertias (e.g., hip/knee vs. ankle/wrist), while retaining per-joint safety constraints through effort/velocity saturation.

\paragraph{Safety limits and action scaling.}
We enforce actuator-specific ffort limits (e.g., hips up to 88--139\,Nm; ankles/waist up to 50\,Nm; shoulders/elbows up to 25\,Nm; wrist pitch/yaw up to 5\,Nm) and velocity limits (typically 20--37\,rad/s depending on joint group). 
To keep commanded target offsets within feasible ranges under these limits, we scale position actions by a joint-wise factor proportional to the ratio between torque limits and stiffness,
$
\Delta q_{\max} \propto 0.25\, \tau_{\max}/K_p,
$
which bounds the implied torques for a given position error and improves stability during high-dynamic tracking.

\subsection{RobotBridge}
RobotBridge is the unified deployment layer of our MOSAIC system, designed to bridge high-level policy algorithms with heterogeneous robot platforms in a plug-and-play manner, similar to RoboJudo~\cite{RoboJuDo}, but with more scalable abilities. Its architecture cleanly separates high-level policy execution from low-level robot-specific interfaces. In practice, control policies are implemented in a platform-agnostic, high-level module, while all hardware-specific commands are handled by a dedicated low-level interface built on the robot’s SDK. This design decouples algorithm logic from device details, allowing the same policy code to drive different robots or simulations with minimal modification. For example, the identical MOSAIC policy pipeline can be executed in MuJoCo simulation or deployed on Unitree G1 hardware without any code change, facilitated by RobotBridge’s unified sim-to-real interface.

RobotBridge supports seamless sim-to-real transfer on the Unitree G1 humanoid~\cite{unitree_g1_web} and also accommodates multi-robot simulation scenarios. It currently integrates support for Unitree’s humanoid platforms (H1 and H1\_2~\cite{unitree_h1_web}) and PND's humanoid platform Adam~\cite{pnd_adam_web} in addition to the G1 robot. This enables evaluation with multiple robots in shared environments, though our experiments primarily focus on a single robot. The framework’s modular design further permits running a wide range of policy pipelines under a common interface. We have integrated diverse control policies including our MOSAIC task policy, teleoperation-based imitation controllers like TWIST~\cite{ze2025twist}, general motion trackers such as GMT~\cite{chen2025gmt}, sim-to-real alignment policies like ASAP, locomotion-specific controllers (e.g. AMO~\cite{li2025amo}), and even dual-agent tracking setups (e.g. a BeyondMimic~\cite{liao2025beyondmimic}-style imitation scenario with a target agent and a follower) within RobotBridge. This flexibility showcases how RobotBridge abstracts the interaction between algorithms and robot embodiment: each policy, regardless of its internal mechanics, interfaces with the robot through the same standardized observation and action API provided by RobotBridge’s environment and controller modules. By handling sensor data streams, action scaling, and communication to the robot’s actuators internally, RobotBridge allows researchers to focus on high-level policy development while ensuring robust execution on real hardware.

Detailed deployment and configuration instructions for RobotBridge (evaluation branch) will be available in the open-source repository’s documentation. By following a configuration-driven approach (built on Hydra~\cite{Yadan2019Hydra}), users can easily switch robot models or control pipelines without code changes, simply by adjusting YAML configs or command-line arguments. This makes RobotBridge a convenient and rigorous platform for evaluating many control policies across simulation and real robots under a unified framework, as required for our MOSAIC system evaluation.
\subsection{Offline Replay}
\subsubsection{Pipeline}
Offline motion replay in MOSAIC is implemented as a lightweight configuration change in RobotBridge. 
Crucially, the motion-loading and preprocessing interface used for replay is identical to the one used during training, which enables a consistent pipeline from learning to deployment. 
As a result, the same reference clips can be (i) visualized and verified in MuJoCo for sim-to-sim qualitative inspection and (ii) executed on the real robot for deployment tests, without modifying the policy code or runtime logic.

RobotBridge’s layered design further decouples the high-level components (policy checkpoint and reference motion selection) from the low-level robot stack (SDK, control loop, safety limits). 
Therefore, switching between different policies or motion clips does not interrupt the low-level controller or require reinitializing the robot (e.g., rehanging the robot or preloading all motions/models before testing). 
In practice, we can iterate over policies and references online while keeping the underlying control and communication pipeline unchanged, facilitating rapid and repeatable evaluation.
\subsection{Online Teleoperation}
\subsubsection{Pipeline}
\begin{figure}[!t]
    \centering
    \includegraphics[width=1.0\linewidth]{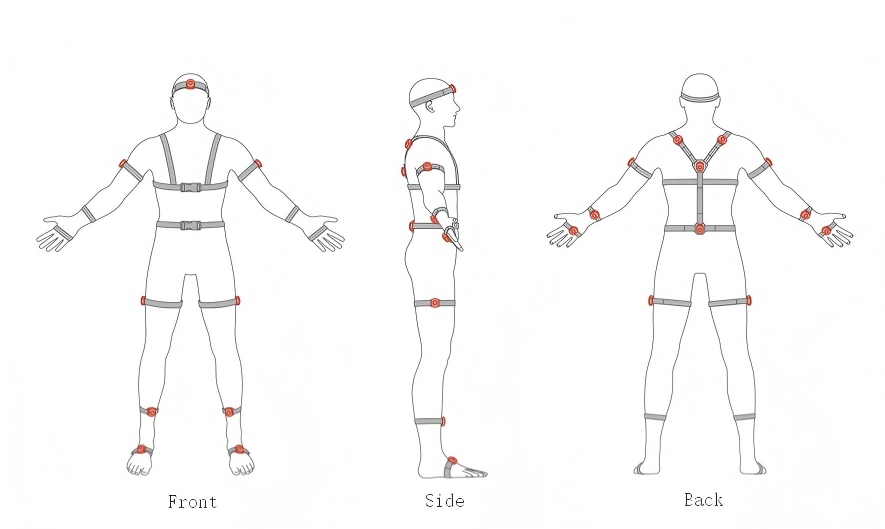}
    \caption{\textbf{Schematic of the Full-body Inertial Tracker Configuration.} To ensure high precision and motion fidelity during data acquisition, we adopt a comprehensive setup consisting of 17 inertial markers.}
    \label{fig:noitom_tracker_layout}
\end{figure}

We developed an online teleoperation pipeline supporting two distinct hardware interfaces: Virtual Reality (VR) and inertial motion capture. The VR system utilizes PICO 4 Ultra, consisting of 5 tracking straps and 2 handheld controllers, while the inertial motion capture system utilizes a Noitom suit with 17 trackers via Noitom Axis Studio \cite{noitom_axis} (see Figure~\ref{fig:noitom_tracker_layout}). Both interfaces utilize a unified processing pipeline to generate the reference motion stream:

\begin{enumerate}
    \item \textbf{Data Transmission and Retargeting:}
    Motion signals from the VR system are transmitted to a local workstation via Wi-Fi, while the Noitom inertial suit is connected via a wired interface. On the workstation, human kinematics are retargeted to the robot's DoF using the GMR framework \cite{ze2025gmr}. The resulting raw joint positions and root quaternions are then streamed to the robot's onboard computer via the Lightweight Communications and Marshalling (LCM)~\cite{lcm_github} protocol.
    
    \item \textbf{Onboard Signal Processing:} 
    Upon receiving the raw data, the robot's onboard host performs real-time processing to ensure signal quality. The joint positions undergo \textit{Exponential Moving Average} (EMA) smoothing to mitigate sensor noise and transmission jitter. Joint velocities are then estimated by computing the gradient over a rolling position buffer using a central difference scheme. Finally, the \textit{anchor orientation} is derived via forward kinematics (FK) from the root orientation and processed joint states, completing the set of reference observations required for real-time policy inference.
\end{enumerate}

\subsubsection{Delay Analysis}
\begin{figure}[!t]
    \centering
    \includegraphics[width=1.0\linewidth]{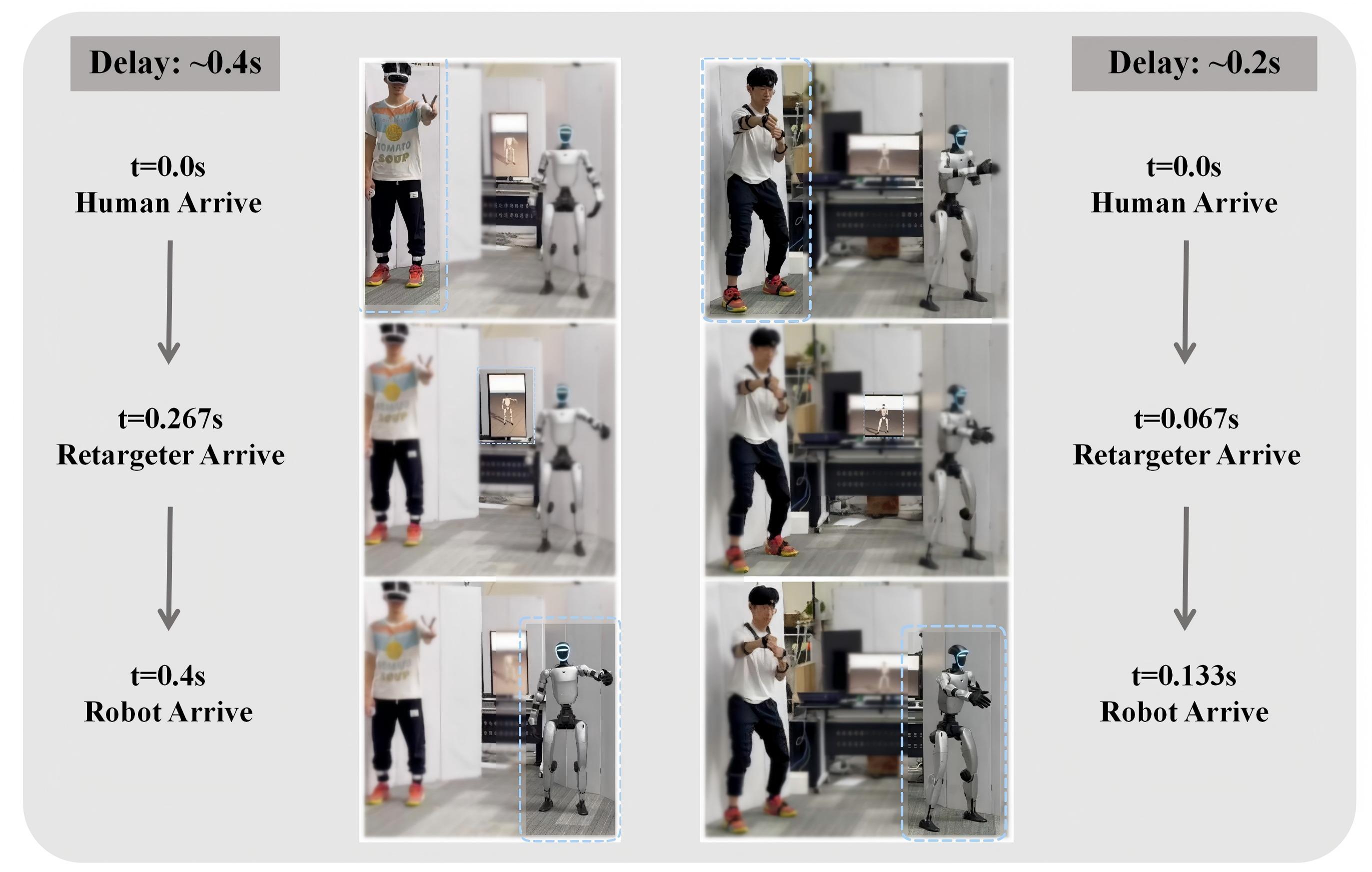}
    \caption{\textbf{Delay Analysis.} The teleoperation delay, as measured via video analysis, is approximately 0.4 s for the VR system and 0.2 s for the inertial motion capture system.}
    \label{fig:delay}
\end{figure}

As illustrated in Figure ~\ref{fig:delay}, the system exhibits an end-to-end teleoperation latency of approximately 0.4 s for the VR-based setup and 0.2 s for the inertial motion capture system averaged over 10 tests. The first phase, comprising data transmission from the device to the local workstation and the subsequent retargeting, accounts for 0.267 s in the VR setup and 0.067 s in the MoCap system, is the main contribution source for the delay. To mitigate the affect of delays for the teleoperation system, we use EMA smoothing and interpolation to the output retargeting results.

\section*{Network Ablations}
\begin{figure}[!t]
    \centering
    \includegraphics[width=1.0\linewidth]{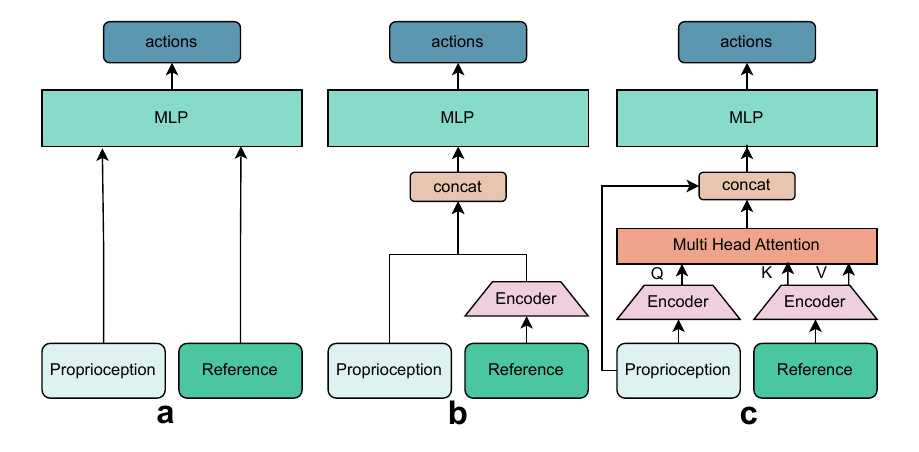}
    \caption{\textbf{Policy Network Ablations for Fusing Proprioception and Reference.}
We compare three architectures with the same input interface (robot proprioception and one-step reference) and the same output (full-body action).
\textbf{(a) MLP fusion:} proprioception and reference are concatenated and directly mapped to actions by an MLP.
\textbf{(b) Encoder-based fusion:} the reference is first encoded into a compact latent (e.g., FSQ or VQ-VAE style code), then concatenated with proprioception and fed to an MLP head.
\textbf{(c) Attention-based fusion:} proprioception features form queries ($Q$) that attend to reference features as keys/values ($K,V$) via multi-head attention; the attended feature is concatenated with proprioception and passed through an MLP to produce actions.}
    \label{fig:net_ablation}
\end{figure}

We ablate three policy architectures for fusing proprioception and the one-step reference input (Shown in Fig.~\ref{fig:net_ablation}): 
(a) MLP fusion (our final choice), where the reference and proprioception are directly concatenated and mapped to actions by a multi-layer perceptron; 
(b) encoder-based fusion, where the reference is first compressed by a learned motion encoder (implemented with FSQ~\cite{mentzer2023fsq} or VQ-VAE~\cite{van2017vqvae} style discretization) and then fused with proprioception via an MLP head; and 
(c) attention-based fusion~\cite{he2025attention}, where proprioception queries attend to reference features via multi-head attention before an MLP output head.

\begin{figure*}[t]
    \centering
    \begin{subfigure}[t]{0.49\linewidth}
        \centering
        \includegraphics[width=\linewidth]{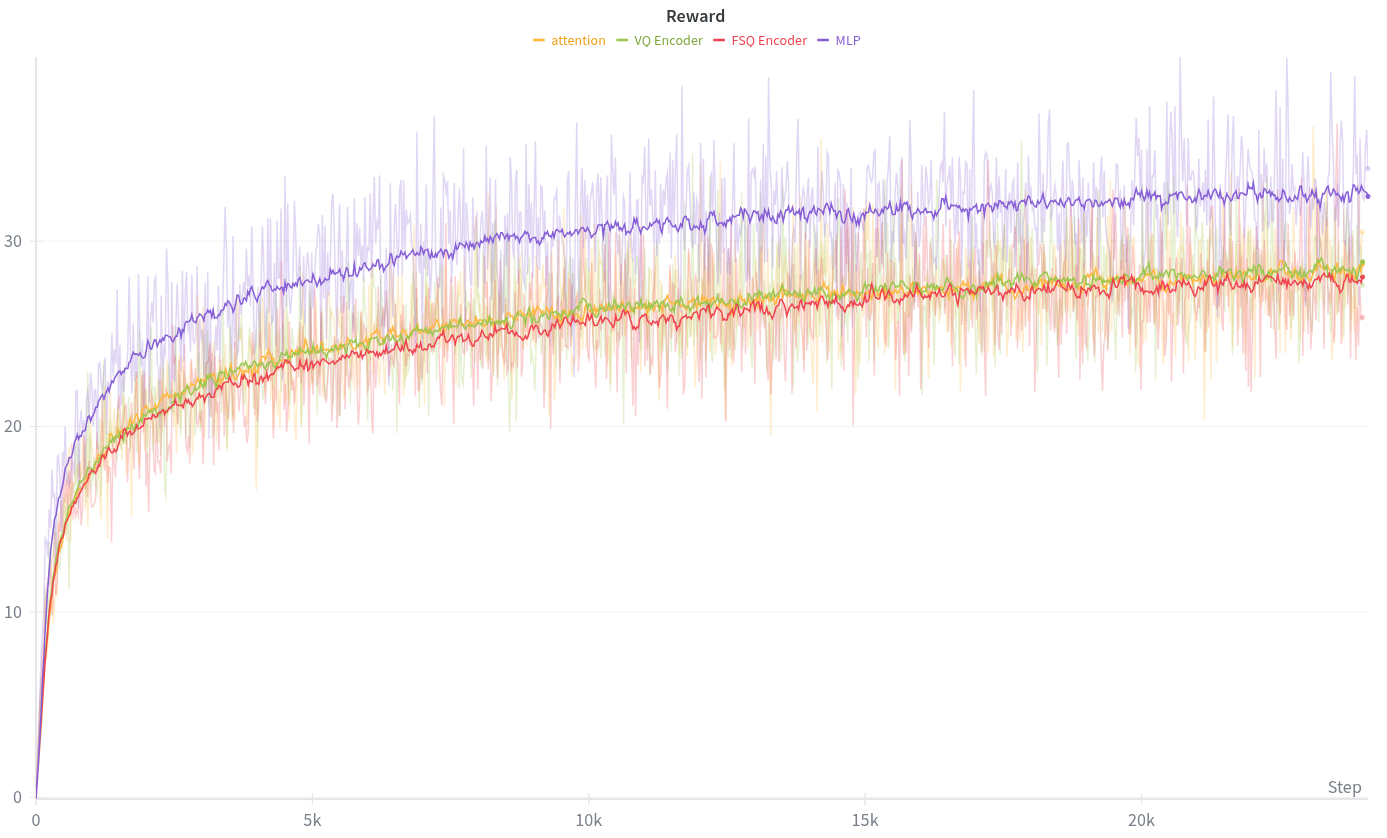}
        \caption{\textbf{Reward.}}
        \label{fig:net_ablation_reward}
    \end{subfigure}\hfill
    \begin{subfigure}[t]{0.49\linewidth}
        \centering
        \includegraphics[width=\linewidth]{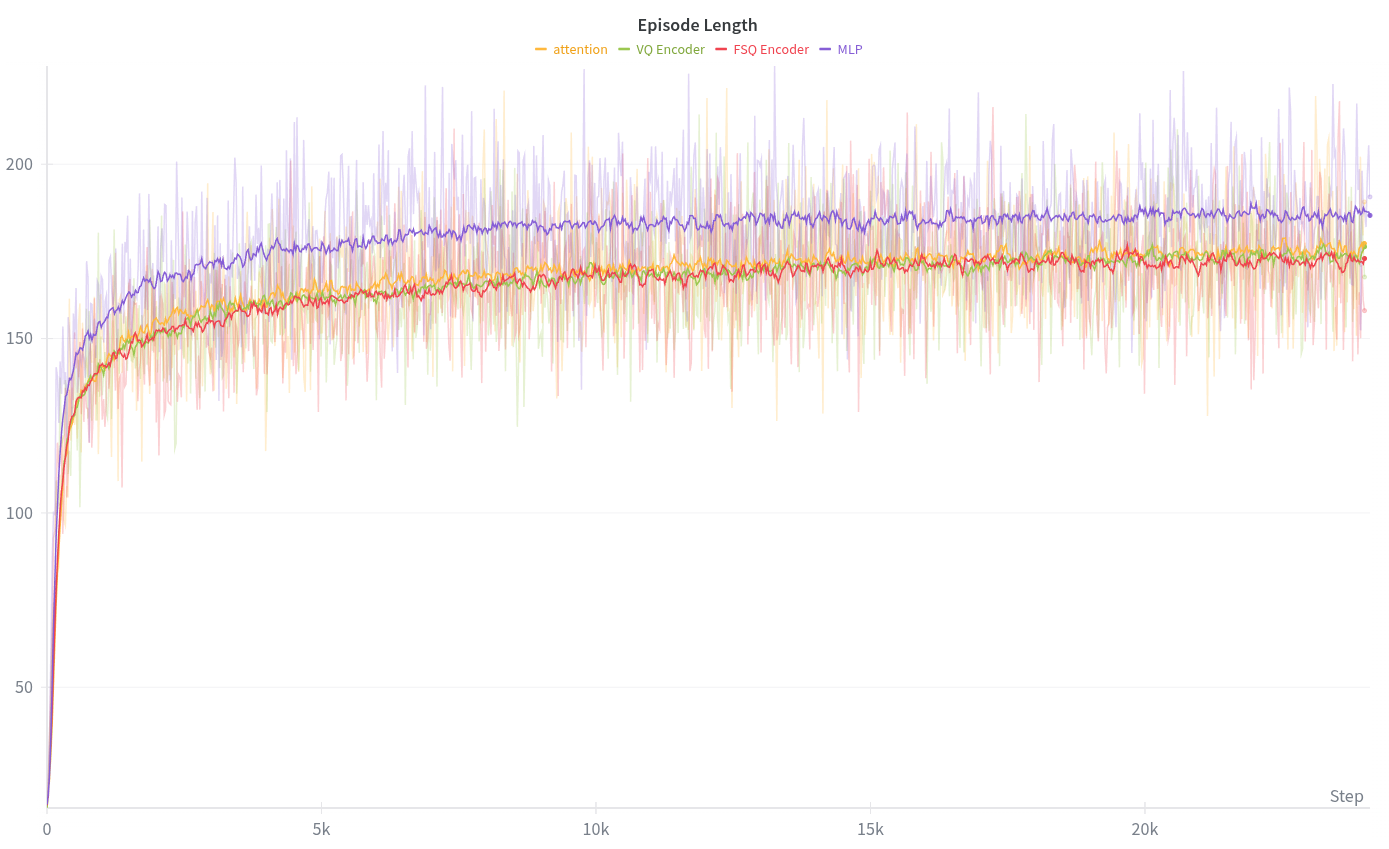}
        \caption{\textbf{Episode length.}}
        \label{fig:net_ablation_ep_len}
    \end{subfigure}
    \caption{\textbf{Network-architecture ablation.} Training curves comparing the MLP actor (ours) against encoder-based actors (VQ, FSQ) and an attention-based actor, under identical PPO settings with the same critic. The MLP achieves higher reward and longer episode length, while more complex actors do not yield gains despite increased training cost.}
    \label{fig:net_ablation_curves}
\end{figure*}

Across training curves shown in Fig.~\ref{fig:net_ablation_curves}, the encoder-based (FSQ/VQ) and attention-based actors do not provide consistent gains over the MLP baseline in either tracking reward or episode length. These variants also incur higher training cost due to additional reconstruction/codebook updates or attention blocks. We therefore adopt MLP fusion and scale capacity via wider layers, which is a more reliable lever than architectural complexity at our current data scale. 
We attribute this to the relatively constrained structure of humanoid motions after retargeting into a unified robot space: despite stylistic diversity, the underlying control regularities are sufficiently captured by a moderately large MLP, and more sophisticated fusion mechanisms provide limited marginal benefit at the current data scale.

In all network ablations, we only change the actor architecture while keeping the critic identical across runs. 
All PPO hyperparameters and training settings are held constant; the only difference is the actor’s fusion mechanism and its associated architectural parameters (Table~\ref{tab:net_ablation_params}).

\begin{table}[t]
\centering
\small
\setlength{\tabcolsep}{4pt}
\renewcommand{\arraystretch}{1.15}
\begin{tabular}{lc}
\toprule
\textbf{Config} & \textbf{Parameters} \\
\midrule
\textbf{FSQ encoder}  & \\
encoder\_hidden\_dims & [512, 128] \\
latent\_dim & 16 \\
num\_levels & 7 \\
mlp\_hidden\_dims & [512, 256] \\
activation\_mlp & ELU \\
activation\_fsq & ELU \\
\midrule
\textbf{VQ encoder} & \\
encoder\_hidden\_dims & [512, 256] \\
encoder\_output\_dim & 128 \\
num\_embeddings & 512 \\
embedding\_dim & 8 \\
commitment\_weight & 0.25 \\
vq\_loss\_coef & 1.0 \\
mlp\_hidden\_dims & [512, 256] \\
activation\_mlp & ELU \\
activation\_vq & ELU \\
\midrule
\textbf{attention} & \\
encoder\_hidden\_dims & [512, 256] \\
attention\_dim & 128 \\
num\_head & 2 \\
mlp\_hidden\_dims & [512, 256] \\
activation\_mlp & ELU \\
activation\_attention & ELU \\
\bottomrule
\end{tabular}
\caption{\textbf{Network ablation configurations.} All runs share the same critic architecture and training hyperparameters; only the actor structure differs.}
\label{tab:net_ablation_params}
\end{table}

\section*{FLD Augmentation}
\label{sec:fld}
\subsection{FLD Training}

The Fourier Latent Dynamics (FLD)~\cite{li2024fld} model was trained on a small curated set of teleoperated periodic locomotion demonstrations, comprising approximately two minutes of data with 30 distinct gait instances.
These demonstrations exhibit stable cyclic structures with diverse step frequencies, step lengths, and operator-dependent styles.

FLD represents periodic motion using a low-dimensional latent state consisting of
a time-varying phase variable $\phi_t$ and quasi time-invariant global style parameters
$\boldsymbol{\theta}$.
Given an initial latent state $(\phi_t, \boldsymbol{\theta})$, future latent states are propagated by explicit phase dynamics,
\begin{equation}
    \phi_{t+k} = \phi_t + k f \Delta t,
\end{equation}
where $f$ denotes the learned gait frequency and $\Delta t$ is the simulation time step.
The corresponding motion state $\hat{\mathbf{s}}_{t+k}$ is decoded via a Fourier-based latent decoder:
\begin{equation}
    \hat{\mathbf{s}}_{t+k} = D\!\left(\phi_{t+k}, \boldsymbol{\theta}\right).
\end{equation}

The model is trained to reconstruct motion trajectories over a multi-step horizon.
Specifically, the reconstruction objective is defined as
\begin{equation}
    \mathcal{L}_{\mathrm{rec}}
    = \sum_{k=0}^{K} \alpha^{k}
    \left\| \hat{\mathbf{s}}_{t+k} - \mathbf{s}_{t+k} \right\|_2^2,
\end{equation}
where $\alpha \in (0,1]$ is a temporal discount factor that emphasizes near-term accuracy while enforcing long-horizon consistency.

To encourage coherent temporal evolution, we further impose a phase-consistency regularization,
which penalizes deviations from monotonic and smooth phase progression:
\begin{equation}
    \mathcal{L}_{\mathrm{phase}}
    = \sum_{t} \left\| (\phi_{t+1} - \phi_t) - f \Delta t \right\|_2^2.
\end{equation}

The final training objective is given by
\begin{equation}
    \mathcal{L}_{\mathrm{FLD}}
    = \mathcal{L}_{\mathrm{rec}} + \lambda_{\phi}\, \mathcal{L}_{\mathrm{phase}},
\end{equation}
where $\lambda_{\phi}$ controls the strength of the phase regularization.

All motion sequences are temporally normalized to a fixed frame rate and segmented into complete gait cycles prior to training.
The model is optimized using Adam until convergence.
No additional supervision on contact states, ground reaction forces, or robot dynamics is introduced, keeping FLD purely self-supervised.

\paragraph{FLD model state space.}
FLD representation learning's state space consists of the base linear velocity, base angular velocity, base height, the projected gravity vector expressed in the robot frame, and joint positions.
All quantities are temporally aligned with the motion trajectories used for FLD training and normalized to a fixed frame rate.

\paragraph{Representation training parameters.}
The FLD model training hyperparameters are summarized in Table~\ref{tab:fld_params}, and the FLD model architecture is listed in Table~\ref{tab:fld_arch}

\subsection{Latent-Space Sampling and Motion Synthesis}
After training, we model the distribution of the learned global style parameters
$\boldsymbol{\theta}$ using a Gaussian Mixture Model (GMM)~\cite{li2024fld}.
This parametric density captures the variability of gait frequency, amplitude,
and stylistic factors observed in the limited set of teleoperated demonstrations.

Novel periodic locomotion trajectories are synthesized by sampling latent parameters
from the learned GMM and initializing the phase uniformly:
\begin{equation}
    \boldsymbol{\theta} \sim p_{\mathrm{GMM}}(\boldsymbol{\theta}),
    \qquad
    \phi_0 \sim \mathcal{U}(0, 2\pi).
\end{equation}
Given the sampled latent state $(\phi_0, \boldsymbol{\theta})$,
future phases are deterministically propagated according to the learned frequency
$f(\boldsymbol{\theta})$:
\begin{equation}
    \phi_{t+k} = \phi_0 + k f(\boldsymbol{\theta}) \Delta t,
\end{equation}
and decoded into motion states using the FLD decoder:
\begin{equation}
    \hat{\mathbf{s}}_{t+k} = D\!\left(\phi_{t+k}, \boldsymbol{\theta}\right).
\end{equation}

By explicitly propagate the phase variable, this procedure generates
long-horizon periodic locomotion trajectories that preserve global gait consistency
over extended time horizons.
In total, we synthesized approximately 10 hours of periodic locomotion motions,
which were used exclusively for ablation studies on dataset augmentation.

The full state composition is summarized in Table~\ref{tab:fld_state}.

\begin{table}[h]
    \centering
    \small
    \setlength{\tabcolsep}{6pt}
    \renewcommand{\arraystretch}{1.2}
    \begin{tabular}{lc}
    \toprule
    \textbf{Entry} & \textbf{Dimensions} \\
    \midrule
    base linear velocity        & 3 \\
    base angular velocity       & 3 \\
    base height                 & 1 \\
    projected gravity           & 3 \\
    joint positions             & 29 \\
    \bottomrule
    \end{tabular}
    \caption{\textbf{State Space Used for FLD Representation Learning.}}
    \label{tab:fld_state}
\end{table}

\begin{table}[h]
    \centering
    \small
    \setlength{\tabcolsep}{6pt}
    \renewcommand{\arraystretch}{1.2}
    \begin{tabular}{lcc}
    \toprule
    \textbf{Parameter} & \textbf{Symbol} & \textbf{Value} \\
    \midrule
    step time (seconds) & $\Delta t$ & 0.02 \\
    learning rate & -- & $1\times10^{-4}$ \\
    weight decay & -- & $5\times10^{-4}$ \\
    learning epochs & -- & 5 \\
    mini-batches & -- & 10 \\
    latent channels & $c$ & 32 \\
    trajectory segment length & $H$ & 91 \\
    FLD propagation horizon & $N$ & 90 \\
    propagation decay & $\alpha$ & 1.0 \\
    \bottomrule
    \end{tabular}
    \caption{\textbf{FLD Representation Training Parameters.}}
    \label{tab:fld_params}
\end{table}

\begin{table}[h]
    \centering
    \small
    \setlength{\tabcolsep}{6pt}
    \renewcommand{\arraystretch}{1.2}
    \begin{tabular}{lcccc}
    \toprule
    \textbf{Module} & \textbf{Layer} & \textbf{Output Size} & \textbf{Activation} \\
    \midrule
    Encoder
        & Conv1D & $256 \times 91$ & ELU \\
        & Conv1D & $256 \times 91$ & ELU \\
        & Conv1D & $8 \times 91$  & ELU \\
    \midrule
    \addlinespace
    \multirow{2}{*}{\makecell[l]{Phase-\\Encoder}}
        & Linear & $32 \times 2$ & Atan2 \\
        &  &  &  &  \\

    \midrule
    \addlinespace
    Decoder
        & Conv1D & $256 \times 91$ & ELU \\
        & Conv1D & $256 \times 91$ & ELU \\
        & Conv1D & $39 \times 91$ & ELU \\
    \bottomrule
    \end{tabular}
    \caption{\textbf{Network Architecture of FLD Model.}}
    \label{tab:fld_arch}
\end{table}
\section*{Open-Source Checklist}
\label{sec:checklist}
To support reproducibility and downstream research, we will release code, models, and data for MOSAIC, together with the full data-processing pipeline and deployment stack.

\paragraph{Code (training).}
We will open-source the MOSAIC training framework, including: (i) PPO training scripts and configuration files, (ii) observation/reward/termination specifications used in all experiments, (iii) the multi-motion data loader and two-level resampling implementation (motion-level and within-motion), (iv) ablation utilities (e.g., RL vs.\ distillation; adaptor variants; network-architecture ablations), and (v) evaluation scripts for simulation benchmarks.

\paragraph{Code (deployment).}
We will open-source RobotBridge, for sim2sim visualization and sim2real testing without changing the pipeline.

\paragraph{Data (redistributable).}
We will release our in-house motion data used by MOSAIC, including: (i) optical MoCap, (ii) inertial MoCap, and (iii) small teleoperation adaptation datasets collected for adaptor training. For each sequence, we will provide two standardized formats: AMASS-style human data and G1-retargeted robot data, together with metadata (source, frame rate, sequence duration, filtering criteria, and splits).

\paragraph{Generated motions.}
We will release (i) the text prompts used for motion generation (covering seven action categories with planar constraints and $\ge$10\,s duration), (ii) the filtered generated motions (AMASS-style + G1-retargeted), and (iii) scripts that reproduce the generation, retargeting, and manual filtering pipeline.

All open-sourced data are with labels, which describe motion behaviors.

\paragraph{Public datasets (scripts only).}
Due to third-party licensing, we will not redistribute AMASS/OMOMO directly; instead we will release scripts to apply the same retargeting/standardization pipeline used for our in-house and generated data (coordinate normalization, initial pose alignment, and unified serialization).

\paragraph{Models and checkpoints.}
We will release trained checkpoints for: (i) the teleoperation-oriented general motion tracker, (ii) interface-specific adaptor policies, and (iii) the residual module used for distillation, along with exact training configurations and seeds required to reproduce the reported results.
\section*{Motion Prompt Examples for Motion Generation}
\label{sec:prompt}
\textbf{Prompt template.}
We follow a simple, reproducible template:
\begin{quote}
\small
\textit{``A person [action description], [style/emotion optional], continuing for at least 10 seconds, on flat ground.''}
\end{quote}

\textbf{Examples.}
\begin{itemize}[leftmargin=1.2em]
    \item \textbf{Static pose:} \textit{A person holding a deep squat position (knees bent, thighs parallel to the floor) and not moving, maintaining it for more than 10 seconds.}
    \item \textbf{Running:} \textit{A person confidently running at a steady pace with an upright posture, continuing for at least 10 seconds.}
    \item \textbf{Walking:} \textit{A person absentmindedly walking as if balancing a book on their head, steps deliberate and body straight, continuing for more than 10 seconds.}
    \item \textbf{Daily activity:} \textit{A person angrily folding imaginary clothes, making aggressive, forceful movements, continuing for over 10 seconds.}
    \item \textbf{Martial arts:} \textit{A person performing a slow tai chi routine, shifting weight gracefully through each pose, continuing for more than 10 seconds.}
    \item \textbf{Dancing:} \textit{A person dancing slowly in a sorrowful manner, moving their body with a somber, expressive style, continuing for more than 10 seconds.}
    \item \textbf{Sports exercise:} \textit{A person doing push-ups rapidly, pushing up and down as fast as possible for 10 seconds, continuing for more than 10 seconds.}
\end{itemize}

\end{document}